\documentclass{article}
\usepackage[final, nonatbib]{neurips_2024}

\usepackage[utf8]{inputenc} 
\usepackage[T1]{fontenc}    
\usepackage{hyperref}       
\usepackage{url}            
\usepackage{booktabs}       
\usepackage{amsfonts}       
\usepackage{nicefrac}       
\usepackage{microtype}      
\usepackage{xcolor}         
\usepackage{makecell}
\usepackage[maxnames=100]{biblatex}
\usepackage{graphicx}
\usepackage{booktabs}
\usepackage{indentfirst}

\usepackage[accsupp]{axessibility}  

\usepackage{amsmath}
\usepackage{amssymb}
\usepackage{float}
\usepackage{enumitem}
\usepackage{multirow}
\usepackage[mathscr]{eucal}
\usepackage[export]{adjustbox}
\usepackage{multicol,lipsum}
\usepackage{arydshln}
\usepackage{pifont}
\usepackage{booktabs}
\usepackage{bbding}

\def\0{{\bf 0}}
\def\1{{\bf 1}}

\definecolor{purple}{rgb}{0.56,0.27,0.68}
\definecolor{red}{rgb}{0.95,0.4,0.4}
\definecolor{purered}{rgb}{1,0,0}
\definecolor{blue}{rgb}{0.4,0.4,0.95}
\definecolor{darkblue}{rgb}{0,0,0.8}
\definecolor{grey}{rgb}{0.6,0.6,0.6}
\definecolor{col1}{RGB}{232, 161, 148}
\definecolor{col11}{RGB}{255, 228, 228}
\definecolor{col2}{RGB}{148, 187, 232}
\definecolor{col33}{RGB}{206, 239, 255}
\definecolor{col3}{RGB}{233, 255, 245}
\definecolor{lightgrey}{rgb}{0.85,0.85,0.85}
\definecolor{lightlightgrey}{rgb}{0.9,0.9,0.9}
\definecolor{verylightBG}{rgb}{0.9,0.99,0.99}
\definecolor{darkgreen}{rgb}{0., 0.85, 0.5}

\definecolor{gtred}{RGB}{204, 0, 0}
\definecolor{predgreen}{RGB}{31, 237, 31}
\definecolor{figGreen}{RGB}{56, 118, 29}

\graphicspath{{./figures/}}   

\newcommand\purered[1]{\textcolor{purered}{#1}}

\usepackage{algorithm2e}
\usepackage{pythonhighlight} 
\usepackage{fontawesome5}
\lstnewenvironment{pythonic}[1][]{\lstset{style=mypython, frame=none, #1}}{}

\usepackage[capitalize]{cleveref}
\crefname{section}{Sec.}{Secs.}
\Crefname{section}{Section}{Sections}
\Crefname{table}{Table}{Tables}
\crefname{table}{Tab.}{Tabs.}

\title{Revisiting Few-Shot Object Detection with Vision-Language Models}

\author{%
Anish Madan$^{1,}\thanks{Equal Contribution}$, Neehar Peri$^{1,*}$, Shu Kong$^{2,3,}\thanks{Equal Senior Authorship}$, Deva Ramanan$^{1,\dagger}$\\
$^1$Carnegie Mellon University, $^2$University of Macau, $^3$Institute of Collaborative Innovation\\
}

\begin{document}
\maketitle

\begin{abstract}
The era of vision-language models (VLMs) trained on web-scale datasets challenges conventional formulations of “open-world" perception. In this work, we revisit the task of few-shot object detection (FSOD) in the context of recent foundational VLMs. First, we point out that zero-shot predictions from VLMs such as GroundingDINO significantly outperform state-of-the-art few-shot detectors (48 vs. 33 AP) on COCO. Despite their strong zero-shot performance, such foundation models may still be sub-optimal. For example, {\tt trucks} on the web may be defined
differently from {\tt trucks} for a target application such as autonomous vehicle perception. We argue that the task of few-shot recognition can be reformulated as aligning foundation models to target concepts using a few examples. Interestingly, such examples can be multi-modal, using both text and visual cues, mimicking
instructions that are often given to human annotators when defining a target concept of interest. Concretely, we propose Foundational FSOD, a new benchmark protocol that evaluates detectors pre-trained on any external data and fine-tuned on multi-modal (text and visual) K-shot examples per target class. We repurpose nuImages for Foundational FSOD, benchmark several popular open-source VLMs, and provide an empirical analysis of state-of-the-art methods. Lastly, we discuss our recent CVPR 2024 Foundational FSOD competition and share insights from the community. Notably, the winning team significantly outperforms our baseline by 23.3 mAP! Our code and dataset splits are available on \href{https://github.com/anishmadan23/foundational_fsod}{GitHub} and \href{https://huggingface.co/anishmadan23/foundational_fsod/tree/main}{HuggingFace}.
\end{abstract}

\section{Introduction}
\label{sec:intro}




Vision-language models (VLMs) trained on (often proprietary) web-scale datasets have disrupted traditional notions of the ``open-world'', particularly for few-shot recognition. In this paper, we revisit few-shot object detection (FSOD) in the context of these foundation models, propose a new benchmark protocol that allows foundation models to ``enter the conversation'', and present several simple baselines. 

First, we highlight that {\em zero-shot} predictions from VLMs like GroundingDINO \cite{liu2023grounding} demonstrate a remarkable improvement over state-of-the-art {\em few-shot} detectors (48.3 vs. 33.1 AP) on COCO \cite{Lin2014MicrosoftCC}, as shown in Table \ref{tab:gdino_zs}. In hindsight, this is not surprising, as the former is pre-trained on far more data (that may include visual examples of the target concept), while the later is pre-trained on data that is explicitly curated to avoid target concepts of interest. From this perspective, VLMs violate the current training protocol of few-shot benchmarks, suggesting that such protocols need to be rethought in the foundational era.

\begin{figure*}[t!]
    \centering
    \includegraphics[width=1.0\linewidth]{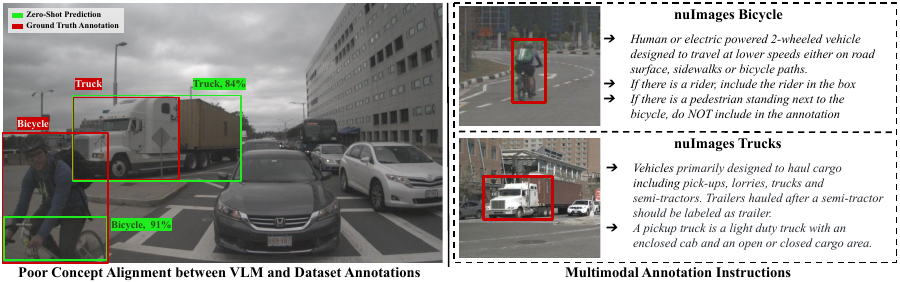}
    \vspace{-3mm}
    \caption{\small
    \textbf{Poor Alignment Between Vision Language Models (VLMs) and Target Concepts}. 
    Although VLMs show impressive zero-shot performance, they struggle when the target class is different from concepts encountered in web-scale training. On the {\bf left}, we see that the nuImages dataset \cite{caesar2020nuscenes}  defines the cab of the {\tt truck} as a separate concept from its {\tt trailer} (shown in \textbf{\textcolor{gtred}{red}}). In contrast, the VLM predicts the entire vehicle as a {\tt truck} (shown in \textbf{\textcolor{predgreen}{green}}). Similarly, nuImages annotations dictate that a person riding a bicycle must also be labeled as part of {\tt bicycle} (shown in \textbf{\textcolor{gtred}{red}}) unlike the VLM prediction (in \textbf{\textcolor{predgreen}{green}}). On the {\bf right}, we present the actual {\em class definitions} given to the \href{https://github.com/nutonomy/nuscenes-devkit/blob/master/docs/instructions_nuimages.md}{nuImages annotators}, provided as both textual descriptions and visual examples. Just as human annotators learn concepts from few-shot multi-modal examples, we argue that VLMs should be aligned with $K$ vision-language examples.}
\vspace{-3mm}
    \label{fig:vlm_gap}
\end{figure*}

{\bf Concept Alignment.} Despite their impressive performance, foundation models used in a zero-shot fashion can still be sub-optimal. 
For example, {\tt trucks} as defined for a particular target application like perception for autonomous vehicles may differ from {\tt trucks} as found on the web (cf. Fig.~\ref{fig:vlm_gap}). Indeed, this well-known observation has created the ad-hoc practice of prompt engineering, where users actively search for a textual prompt that elicits the desired zero-shot behaviour. Instead, we argue that one can principally address 
the challenge of {\em aligning} foundation models to target concepts through the lens of few-shot recognition, by presenting VLMs with a few examples of the target concept. Crucially, such examples can be multi-modal, using both text and visual cues, mimicking the natural few-shot {\em multi-modal instructions} that are often given to human annotators when defining a target concept of interest~\cite{chang2023thinking}. Before introducing our new protocol, we first review the conventional FSOD setup below.

{\bf Conventional FSOD.} Existing FSOD benchmarks partition object detection datasets like PASCAL VOC \cite{Everingham10} and COCO \cite{Lin2014MicrosoftCC} into {\tt base} and {\tt novel} classes.  Detectors pre-train on {\tt base} classes and then learn to identify {\tt novel} classes given $K$ examples (or $K$-shots). Current protocols enforce {\tt base} and {\tt novel} to be disjoint to prevent concept leakage, allowing one to evaluate generalization to the ``open-world". However, as most detectors are pre-trained on ImageNet \cite{imagenet}, we point out that {\em concept leakage already occurs in the current FSOD protocol}. For example, {\tt cat} and {\tt person} are deemed {\tt novel} for COCO-FSOD  but are present in ImageNet data used to pre-train detectors \cite{wang2020frustratingly}. Moreoever, {\tt car} is deemed {\tt novel}, but similar concepts like {\tt sports car} and {\tt race car} are present in ImageNet, illustrating the difficulty of even defining leakage. 

\begin{figure}[t]
    \centering 
     \includegraphics[width=0.99\linewidth]{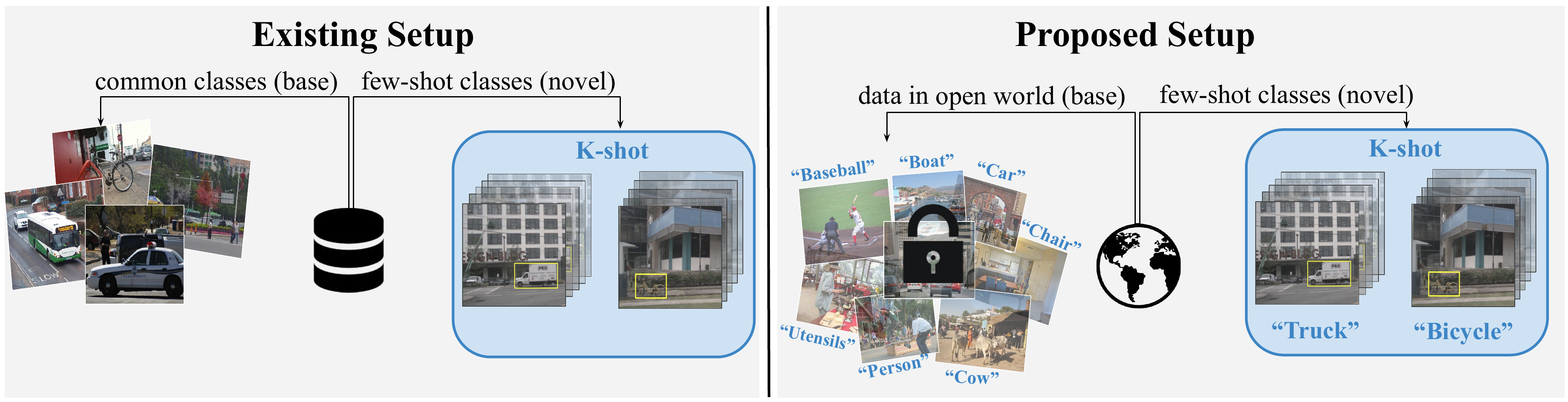}
    \vspace{-1mm}
    \caption{\small
    {\bf Foundational Few-Shot Object Detection (FSOD)}. Conventional FSOD protocols ({\bf left}) allow for pre-training on {\tt base} classes (with many examples per class) and then fine-tuning on $K$-shots of {\tt novel} classes, where {\tt novel} and {\tt base} are designed to be disjoint. 
    However, we point out that pre-training datasets such as ImageNet often contain classes similar to {\tt novel} classes, highlighting the issue of concept leakage.
    As preventing concept leakage is difficult (if not impossible) and appears to be  artificial in the foundational era, we propose  \emph{Foundational FSOD} (\textbf{right}).
    Our setup allows for pre-training on massive (and potentially proprietary) datasets, typical for foundational vision-language models.
    Since these models can process both text and images, one can utilize such {\em multi-modal} $K$-shot examples to {\em align} VLMs with the target concepts of interest.
    }
    \label{fig:fsod}
    \vspace{-2mm}
\end{figure}

{\bf Foundational FSOD.}  We believe that concept leakage should be embraced.  
Our Foundational FSOD protocol replaces the {\tt base} pre-training stage with web-scale pre-training, where such data may be proprietary and not fully disclosed \cite{radford2021learning}. {\em We argue that pre-training on large-scale data will be the key enabler for generalization to the open world}. Note that this hypothesis is difficult to even evaluate under the conventional few-shot protocol, motivating our setup.
Moreover, another key property is that $K$-shot instances may include multi-modal examples spanning both images and text, motivating a multi-modal adaptation stage that aligns the VLM to target concepts 
 (cf. Fig.~\ref{fig:fsod}). We repurpose nuImages \cite{caesar2020nuscenes}, a challenging dataset due to open-world categories such as {\tt debris} and {\tt pushable-pullable}, for our Foundational FSOD benchmark.

We present three major contributions. 
\begin{itemize}
    \item We modernize FSOD benchmarks by embracing foundational VLMs that are pretrained on internet-scale data. We highlight the practical challenge of using multi-modal few-shot examples to define target semantic concepts (as shown in Fig.~\ref{fig:vlm_gap}).
    \item We adapt nuImages for Foundational FSOD, evaluate popular open-source VLMs, and present an empirical analysis of leading methods.
    \item We highlight the results from our recent \href{https://eval.ai/web/challenges/challenge-page/2270/evaluation}{CVPR 2024 challenge} hosted in conjunction with the \href{https://vplow.github.io/vplow_4th.html}{Workshop on Visual Perception via Learning in An Open World}. 
\end{itemize} 

\section{Related Works}

\textbf{Few-Shot Object Detection} aims to detect new categories with limited training data \cite{kohler2021few}. Recent work explores two primary approaches: meta-learning and transfer learning. Meta-learning-based methods focus on acquiring generalizable features from a set of {\tt base} classes, which can then be applied to identify {\tt novel} classes. 
For example, Kang et al. \cite{kang2019few}  proposes a technique that re-weights features from {\tt base} classes to predict {\tt novel} classes.  
Xiao et al. \cite{xiao2022few} propose a framework addressing both few-shot object detection and few-shot viewpoint estimation.  Fan et al. \cite{fan2020few} introduces a general FSOD network that learns a matching metric between image pairs, while Wu et al. \cite{wu2021universal} enhances object features using a universal prototype. More recently, Xu et al. \cite{xu2023generating} propose a generative approach that is robust to noisy object proposals for {\tt novel} classes. In contrast, transfer learning involves partially freezing network weights pretrained on a {\tt base} dataset to improve a model's ability to generalize to {\tt novel} classes with limited data. Transfer learning approaches often follow a two-stage fine-tuning strategy: first train on {\tt base} classes and then fine-tune the box classifier and regressor with $K$-shots from {\tt novel} classes. This strategy has historically outperformed meta-learning approaches \cite{wang2020frustratingly}. Recent work has primarily focused on improving classification performance. Sun et al. \cite{sun2021fsce} utilizes a contrastive proposal encoding loss to encourage instance-level intra-class compactness and inter-class variance. Similarly, Li et al. \cite{li2021beyond} applies a class margin loss to balance inter and intra-class margins. 

\textbf{Vision Language Models} are trained on a large-scale collection of weakly-supervised image-text pairs collected from the web. These models embed images and text into a shared space, enabling open-vocabulary detection. Early works adapt VLMs for object detection by either distilling the model's predictions for specific image regions \cite{gu2021vild, gu2021open} or directly incorporating detection components into frozen \cite{kuo2022fvlm} or fine-tuned \cite{minderer2022owlvit, minderer2023owlvit2, du2022learning} encoders. In contrast, RegionCLIP \cite{zhong2022regionclip} employs a multi-stage training approach, which involves generating pseudo-labels from captioning data, conducting region-text contrastive pre-training, and fine-tuning on detection data. GLIP \cite{li2021grounded} uses a single text query for the entire image and frames detection as a phrase grounding problem. 
More recently, Detic \cite{zhou2022detecting} addresses long-tail detection performance by leveraging image-level supervision. In the context of open-vocabulary detection, there may be some overlap between categories seen during training and testing. We use the term ``zero-shot'' when a model has never been trained on the target dataset. 

\textbf{Fine-Tuning Foundation Models} is of significant interest across many application areas \cite{hu2021lora, zhang2023adding, gao2024clip}. Standard fine-tuning procedures employ both linear probing \cite{chen2020simple, he2022masked, he2020momentum} and full-finetuning \cite{wang2017growing, wu2023revisiting, kirkpatrick2017overcoming, liu2024few} to adapt models to downstream tasks. However, such methods may be suboptimal as they can be computationally expensive. Instead, recent works like CLIP-Adapter \cite{gao2024clip} and Tip-Adapter \cite{zhang2021tip} fine-tune CLIP using parameter-efficient methods \cite{houlsby2019parameter, zhang2020side, jia2022visual} which optimize lightweight MLPs while keeping the encoder frozen. Similarly, inspired by the success of prefix-tuning in language models \cite{deng2022rlprompt,jiang2020can, haviv2021bertese,gao2020making}, prompt adaptation \cite{lu2022prompt, zhu2023prompt, xing2023dual, zhou2022conditional} replaces hand-crafted prompts like "a photo of a {\tt \{cls\}}" with learned word embeddings. CoOp \cite{zhou2022learning} and other prompting methods \cite{lu2022prompt, zhu2023prompt, zhou2022conditional} finetune CLIP via prefix-tuning. Different from most prior work, we investigate fine-tuning strategies for VLM-based detectors using few-shot {\em multi-modal} examples.



\section{Foundational FSOD Benchmark}
As shown in Fig \ref{fig:fsod}, our proposed Foundational FSOD benchmark utilizes vision-language models (VLMs) pre-trained on diverse, large-scale datasets, which are then aligned to $K$ examples of each target class. We contrast our proposed setup with standard benchmarks and present simple strategies for fine-tuning VLMs below.

\subsection{Foundational FSOD Benchmark}
Existing FSOD benchmarks repurpose well-established datasets like PASCAL VOC \cite{Everingham10} and COCO \cite{Lin2014MicrosoftCC} by partitioning them into {\tt base} and {\tt novel} classes for pre-training and fine-tuning, respectively. For COCO, the 60 categories disjoint with PASCAL VOC are used as {\tt base} classes and the remaining
20 are used as {\tt novel} classes \cite{wang2020frustratingly}. However, this setup is artificial and does not reflect how FSOD is deployed in practice. First, FSOD benchmarks construct a set of {\tt novel} classes that include common concepts such as {\tt car} and {\tt person}, and require FSOD methods to detect these common classes by assuming they have only few examples.
Importantly, VLMs like GroundingDINO \cite{liu2023grounding} can already detect common categories with high accuracy on COCO {\em without fine-tuning} (cf. Table \ref{tab:gdino_zs}). Therefore, we focus on benchmarking Foundational FSOD on more realistic and challenging datasets like nuImages \cite{caesar2020nuscenes}. 
Second, existing FSOD benchmarks require that datasets are partitioned into {\tt base} and {\tt novel} classes, which is infeasible for large-scale (often private) foundational datasets. For example, although CLIP's \cite{radford2021learning} model weights are publicly available, its pre-training dataset is not. Instead, FSOD methods should only fine-tune VLMs on $K$-shot annotations for $C$ target classes (or {\tt novel}, as termed in conventional FSOD benchmarks), and also evaluate performance on these $C$ classes.

\subsection{Few-Shot Multi-Modal Concept Alignment} 
Although VLMs achieve strong zero-shot performance on common classes, they struggle when the target class is different from concepts encountered on the web (cf. Fig. \ref{fig:vlm_gap}). For 
example, nuImages \cite{caesar2020nuscenes} defines the cab of a {\tt truck} as a separate concept from its {\tt trailer}. However, Detic \cite{zhou2022detecting} detects the entire vehicle as {\tt truck}. This fine-grained distinction is provided to human annotators with visual examples and textual descriptions. We explore seven methods for alignment (either explicitly by updating model weights via gradient-based fine-tuning or in-context via prompting) below. 


\textbf{Prompt Engineering} uses expressive descriptions, attributes \cite{menon2022visual}, or synonyms \cite{parashar2023prompting, parashar2024neglected} in the text prompt to manually improve the alignment of foundation model outputs to target concepts of interest. In our case, we prompt VLMs with synonyms of the nuImages classes to improve detection accuracy. For example, we augment the language query for {\tt pushable-pullable} with synonyms like {\tt cart} and {\tt wheel barrow}. We provide a full list of synonyms in Table \ref{tab:cls_syn}.

\textbf{Standard Fine-Tuning} updates the last few layers of a model to adapt to new target classes. For two-stage object detectors, this typically requires training the box regression and classifier head. 
However, we find that standard fine-tuning is sub-optimal, motivating our proposed approach below.

\textbf{Federated Fine-Tuning} leverages a simple but evidently underappreciated observation: few-shot object detection datasets are actually federated datasets \cite{gupta2019lvis}. A federated dataset is comprised of smaller mini-datasets, where each mini-dataset is exhaustively annotated for only a single category. For example, {\tt cars} may or may not appear in the background of the $K$ images annotated with {\tt motorcycles}. However, existing FSOD methods incorrectly assume that no {\tt cars} are present in the background of non-{\tt car} images. We devise a simple loss that incorporates this insight, discussed further in the supplement.

\textbf{Language Prompt Tuning} is an established parameter-efficient strategy  \cite{shin2020autoprompt, lester2021power} for updating text embeddings with few-shot examples via fine-tuning. Concretely, for a given language query (e.g. {\tt stroller}), we first extract a text embedding $P^0$ and only fine-tune the text embedding \cite{li2021grounded}.

\textbf{Visual Prompting} uses images of target concepts that are difficult to describe through text as prompts to learn novel concepts in-context. For example, although {\tt debris} is a difficult catchall category to define through text, we can use image examples to improve concept alignment. Typically, visual prompts are tokenized and fed as inputs to a frozen VLM. 

\textbf{Multi-Modal Prompting} combines language and visual prompting to leverage multi-modal features. Intuitively, multi-modal cues can yield better alignment than uni-modal cues alone; in the above case, ambiguous concepts such as {\tt debris} can be clarified with both textual descriptions (e.g {\tt trash can} and {\tt tree branch}) and visual examples (similar to the multi-modal annotator instructions in Fig.~\ref{fig:vlm_gap}!). Here, visual and language prompts are tokenized and separately fed as inputs to a frozen VLM. Specifically, MQDet \cite{xu2024multi} introduces a lightweight Gated Class Scalable Perceiver module that fuses visual cues and text descriptions in the text encoder via class-wise cross attention layers. 

\textbf{Multi-Modal Chat Assistants} can accomplish many of the same tasks as multi-modal prompting through a multi-modal turn-by-turn conversational interface. However, unlike multi-modal prompting, conversational interfaces allow users to iteratively refine concept definitions, similar to how human annotators often require several rounds of feedback to fully understand the target concept.

\section{Experiments}

We conduct extensive experiments to validate that zero-shot inference from VLMs significantly improves over state-of-the-art FSOD approaches,
suggesting that existing benchmarks should be re-framed to allow foundation models to ``enter the conversation''. 
Moreover, we demonstrate that leveraging language cues, especially those available for free (e.g., class names), are crucial to improving performance on data-constrained tasks like FSOD.
    
{\bf Datasets and Metrics.}
We repurpose nuImages \cite{caesar2020nuscenes} to support the study of Foundational FSOD. This dataset annotates 18 classes, which are divided into groups with {\tt many}, {\tt medium}, and {\tt few} examples~\cite{peri2023towards, ma2023longtailed}. We report average precision (AP) for each cohort. Although this dataset is not traditionally used for FSOD, nuImages' open-world categories like {\tt debris} and {\tt pushable-pullable} make it particularly challenging (even for VLMs), and is a realistic benchmark for Foundational FSOD. We follow the $K$-shot dataset creation process established by \cite{wang2020frustratingly}, described below. 
To construct a $K$-shot dataset, we select a target class $c$ and an image at random. If the total annotations for class $c$ in the image are less than or equal to $K$, we add the image to our dataset. We repeat this process for all classes until we have exactly $K$ annotations per class. Since the specific $K$ examples can have a significant impact on the overall performance, we run each experiment over three random data splits and report the average.

\begin{table}[t]
\centering
\caption{\small 
{\bf VLM Zero-Shot Inference Is a Strong FSOD Baseline.} 
Zero-shot inference with VLMs like GroundingDINO resoundingly outperforms state-of-the-art FSOD methods on the COCO FSOD benchmark, motivating the need to re-frame FSOD to embrace foundation models.
}
\vspace{-2mm}
\def\arraystretch{1.0}%
\resizebox{0.99\linewidth}{!}{
    \begin{tabular}{@{}l@{\ \ \ \ \ \ \ \ \ \ \ \ \ \ \ \ \ \ \ \ \ \ \ \ \ \ \ \ }c@{\ \ \ \ \ \ \ \ \ \ \ \ \ \ \ \ \ \ \ \ \ \ \ \ }c@{\ \ \ \ \ \ \ \ \ \ \ \ \ \ \ \ \ \ \ \ }c@{}}
        \toprule
        \multirow{2}{*}{Approach} & \multicolumn{3}{c}{30-shots}  \\
                            & AP    & Base AP   & Novel AP   \\ 
        \midrule
        FRCN-ft-full~\cite{yan2019meta} & 18.6 & 20.6 & 12.5 \\
        FRCN-BCE~\cite{yan2019meta} & 30.2 & 36.8 & 10.3 \\
        TFA w/ fc~\cite{wang2020frustratingly} & 29.3 & 34.5 & 13.5 \\
        TFA w/cos~\cite{wang2020frustratingly} & 29.9 & 35.3 & 13.6 \\
        MPSR~\cite{wu2020multi} & 17.1 & 18.1 & 14.1 \\
        Meta-RCNN~\cite{yan2019meta} & 7.8 & 7.1 & 9.1 \\
        FsDetView~\cite{xiao2022few} & 10.0 & 9.3 & 12.0 \\
        Retentive R-CNN~\cite{fan2021generalized} & 32.9 & 39.3 & 13.8 \\
        DiGeo~\cite{ma2023digeo} & 33.1 & 39.4 & 14.2 \\ 
        \midrule
        
        \textbf{GroundingDINO (Zero-Shot)}~\cite{liu2023grounding} & \textbf{48.3} & \textbf{46.3} & \textbf{54.3} \\
        \bottomrule
    \end{tabular}
}
\vspace{-1mm}

\label{tab:gdino_zs}
\end{table}

\subsection{Zero-Shot Inference Is A Strong FSOD Baseline}
        
We compare state-of-the-art FSOD methods with zero-shot inference from GroundingDINO \cite{liu2023grounding} on COCO in Table~\ref{tab:gdino_zs}. Surprisingly, GroundingDINO outperforms DiGeo \cite{ma2023digeo} by 16.2\% AP averaged across both {\tt base} and {\tt novel} categories despite never being trained on COCO images.
GroundingDINO's impressive performance is due to its large-scale multi-modal pre-training on Objects365~\cite{shao2019objects365}, GoldG \cite{goldg} and Cap4M~\cite{li2021grounded}.
It is worth noting that GroundingDINO achieves higher AP on {\tt novel} classes than {\tt  base}, suggesting that {\tt  novel} classes in existing benchmarks (e.g., {\tt car} and {\tt person}) are actually not rare in the real world.
Therefore, FSOD benchmarks should be re-framed to reflect real-world applications, motivating our setup.

        \begin{table}[t]
            \centering
            \caption{\small
            {\bf Impact of Large-Scale Pre-Training and Language.} We repurpose nuImages for FSOD following the dataset creation process established by \cite{wang2020frustratingly}. 
            We group categories by frequency into {\tt many}, {\tt medium} and {\tt few} examples per class 
            \cite{peri2023towards, ma2023longtailed}. 
            We fine-tune TFA on $K$ examples, but find low performance, 
            $< 3 $AP. However, by simply pre-training on more data ({\tt LVIS}, {\tt COCO} and ImageNet-21K) and leveraging language cues via a CLIP classifier, 5-shot performance improves from 2.02 AP to 15.12 AP. However, rare (or {\tt few}) classes 
            like {\tt strollers}, {\tt pushable-pullable}, and {\tt debris} remain challenging, motivating our task of VLM alignment. 
        }
        \vspace{-2mm}
            \def\arraystretch{1.0}%
            \resizebox{0.95\linewidth}{!}{
                \begin{tabular}{@{}l@{\ \ \ \ \ \ \ \ \ \ \ \ \ \ \ \ \ \ \ \ \ \ \ }c@{\ \ \ \ \ \ \ \ \ \ \ \ \ \ \ }c@{\ \ \ \ \ \ \ \ \ \ \ \ \ \ \ }c@{\ \ \ \ \ \ \ \ \ \ \ \ \ \ \ }c@{}}
                    \toprule
                    \multirow{2}{*}{Approach} & \multicolumn{4}{c}{Average Precision (AP)}  \\
                                        & {\tt All} &   {\tt Many}    & {\tt Medium}   & {\tt Few}    \\ 
                    \midrule
                    \textbf{5-shot} \\
                    \midrule
                        \quad TFA \cite{wang2020frustratingly} w/ {\tt COCO-base} &  1.33 &	2.78 &	1.43 &	0.23 \\
                        \quad TFA \cite{wang2020frustratingly} w/ {\tt LVIS-base} & 2.02 & 1.69 & 4.08 & 0.58 \\
                        \quad TFA \cite{wang2020frustratingly} w/ {\tt LVIS,IN-21K,} & \multirow{2}{*}{\textbf{15.12}} & \multirow{2}{*}{\textbf{22.74}} & \multirow{2}{*}{\textbf{18.99}} & \multirow{2}{*}{\textbf{4.25}}\\
                        \quad {\tt COCO} + CLIP Classifier & & & & \\
                    \midrule
                    \textbf{10-shot} \\
                    \midrule
                        \quad TFA \cite{wang2020frustratingly} w/ {\tt COCO-base} & 1.21 & 2.55 & 1.19 & 0.31 \\
                        \quad TFA \cite{wang2020frustratingly} w/ {\tt LVIS-base} & 2.27 & 2.05 &	4.51 & 0.58 \\ 
                        \quad TFA \cite{wang2020frustratingly} w/ {\tt LVIS,IN-21K,} & \multirow{2}{*}{\textbf{16.09}} & \multirow{2}{*}{\textbf{25.46}} & \multirow{2}{*}{\textbf{20.00}} & \multirow{2}{*}{\textbf{3.73}}\\
                        \quad {\tt COCO} + CLIP Classifier & & & & \\
                    \midrule
                    \textbf{30-shot} \\
                    \midrule
                        \quad TFA \cite{wang2020frustratingly} w/ {\tt COCO-base} & 1.14 & 2.81 & 0.84 & 0.23\\
                        \quad TFA \cite{wang2020frustratingly} w/ {\tt LVIS-base} & 2.23 &	1.48 &	4.98 &	0.45 \\ 
                        \quad TFA \cite{wang2020frustratingly} w/ {\tt LVIS,IN-21K,} & \multirow{2}{*}{\textbf{17.22}} & \multirow{2}{*}{\textbf{25.98}} & \multirow{2}{*}{\textbf{21.64}} & \multirow{2}{*}{\textbf{4.78}}\\
                        \quad {\tt COCO} + CLIP Classifier & & & & \\
                    \bottomrule
                \end{tabular}
                }
           \label{tab:nuimages_ft}
        \end{table}

\subsection{Foundational FSOD with nuImages}
\label{sec: nuimages_proposed_setup}


In the context of foundational models, we argue that partitioning datasets into {\tt base} and {\tt novel} classes no longer makes sense. Instead, FSOD methods should only fine-tune on $K$-shot annotations for $C$ target classes, and also evaluate performance on these $C$ classes. We pre-train TFA \cite{wang2020frustratingly} on diverse datasets and fine-tune on $K$ examples per class and highlight model performance in Table \ref{tab:nuimages_ft}. 
   We train two variants of TFA trained on {\tt COCO-base} and {\tt LVIS-base} and fine-tune both models on $K$ examples of the nuImages classes. Surprisingly, both variants of TFA achieve less than $3$ AP (cf. Table \ref{tab:nuimages_ft}).  We posit that this is largely due to poor classification performance. Since both LVIS and COCO classes do not significantly overlap with nuImages classes, learning a classifier from few examples is extremely difficult. However, we find that simply re-training TFA with a frozen CLIP-based classifier (similar to Detic) dramatically increases performance, reiterating the utility of language  and web-scale pre-training in data-constrained settings.

\begin{table}[t]
            \centering
            \caption{\small
            \small {\bf Empirical Analysis of Baselines (10-Shot) on our Benchmark}. We evaluate popular VLMs on the nuImages FSOD Benchmark and find that MQ-GLIP performs the best among all baseline models. Notably, it achieves 17.0 mAP zero-shot language-only performance, and achieves 21.4 mAP via zero-shot multi-modal prompting averaged over all classes. We can iteratively prompt GPT-4o for synonyms to describe each of the few-shot examples to expand MQ GLIP's text prompts, further improving performance by 0.6\%. Remarkably, our 2024 competition winners further improved performance to 45.4 mAP, beating our best baseline by 23.3\%.}
            \vspace{-2mm}
            \def\arraystretch{1.0}%
            \resizebox{0.99\linewidth}{!}{
            \setlength{\extrarowheight}{2pt}
                \begin{tabular}{@{}l@{\ \ \ \ \ \ \ \ \ \ \ \ \ \ \ }l@{\ \ \ \ \ \ \ \ \ }c@{\ \ \ \ \ \ \ \ \ \ \ \ \ }c@{\ \ \ \ \ \ \ \ \ }c@{\ \ \ \ \ \ \ \ \ }c@{\ \ \ \ \ \ \ \ \ }c@{}}
                    \toprule
                    \multirow{2}{*}{Approach} & \multirow{2}{*}{Backbone} & 
                    \multirow{2}{*}{Pre-Train Data} & \multicolumn{4}{c}{Average Precision (AP)}  \\
                    \cmidrule(l){4-7} 
                      & &  
                                        & {\tt All} & {\tt Many}    & {\tt Med}   & {\tt Few}    \\ 
                   
                    \midrule
                        \textbf{Zero-Shot Detection} & & & & & & \\
                    \midrule
                        \quad RegionCLIP \cite{zhong2022regionclip}  & RN50 & CC3M & 2.50 &	3.20 &	3.80 &	0.40 \\
                       
                        \quad Detic \cite{zhou2022detecting}  & SWIN-B & LVIS, COCO, IN-21K & 14.40	&25.83	&16.59	&2.32  \\
                    
                        \quad GroundingDINO \cite{liu2023grounding}  & SWIN-T & Objects365, GoldG, Cap4M  & 12.05	& 17.29	& 15.45 &	3.72\\
                        \quad GLIP \cite{li2021grounded}  & SWIN-L & FourODs,GoldG,Cap24M  & 17.01	&23.36	&19.86	&8.40\\
                       
                        \quad MQ-GLIP-Text \cite{xu2024multi}  & SWIN-L & Objects365, FourODs, GoldG, Cap24M &17.01	&23.36	&19.85	&8.41\\
                    \midrule
                    \midrule
                    \textbf{Prompt Engineering} & & &  & & & \\
                    \midrule
                        \quad Detic \cite{zhou2022detecting}  & SWIN-B & LVIS, COCO, IN-21K &14.92	&26.48	&17.29	&2.53\\
                        \quad GLIP \cite{li2021grounded}  & SWIN-L & FourODs, GoldG, Cap24M &17.15	&23.82	&19.36	&9.02\\
                    \midrule
                    \textbf{Standard Fine-Tuning} & & &  & & & \\
                    \midrule
                        \quad RegionCLIP \cite{zhong2022regionclip}  & RN50 & CC3M &3.86	&6.08	&5.13	&0.54\\
                        \quad Detic \cite{zhou2022detecting}  & SWIN-B & LVIS, COCO, IN-21K &16.09	&25.46	&20	&3.73\\
                    \midrule
                    \textbf{Federated Fine-Tuning (Ours)} & &  & & & &\\
                    \midrule
                        \quad Detic \cite {zhou2022detecting}  & SWIN-B & LVIS, COCO, IN-21K &17.24	&28.07	&20.71	&4.18\\
                        \quad Detic \cite {zhou2022detecting} w/ Prompt Engineering  & SWIN-B & LVIS, COCO, IN-21K &17.71	&28.46	&21.14	&4.75\\
                    \midrule
                    \textbf{Language Prompt Tuning} & & &  & & &\\
                    \midrule
                        \quad GLIP \cite{li2021grounded}  & SWIN-L & FourODs,GoldG,Cap24M &19.41	&22.18	&\textbf{25.16}	&\textbf{10.39}\\
                    \midrule
                    \textbf{Visual Prompting} & & &  & & & \\
                    \midrule
                        \quad MQ-GLIP-Image \cite{xu2024multi}  & SWIN-L & Objects365,FourODs,GoldG,Cap24M &14.07	&24.39	&15.89	&3.34\\
                    \midrule
                    \textbf{Multi-Modal Prompting} & &  & & & & \\
                    \midrule
                        \quad MQ-GLIP \cite{xu2024multi}  & SWIN-L & Objects365,FourODs,GoldG,Cap24M &21.42	&32.19	& 23.29	& 10.26\\
                    \midrule
                    \textbf{Multi-Modal Chat Assistants} & & &  & & &\\
                    \midrule
                    \quad GPT-4o Zero-Shot Classification \cite{achiam2023gpt} & \textit{Private} & \textit{Private} &9.95	&16.81	&12.11	&1.71\\
                    \quad MQ-GLIP Iterative Prompting& \emph{Private} & \emph{Private} & \textbf{22.03} & \textbf{33.42} & 24.72 & 9.41 \\
                    \midrule
                    \midrule
                    \textbf{CVPR 2024 Competition Results} & & &  & & &\\
                    \midrule
                    \quad PHP\_hhh & \emph{Private}  & \emph{Private} & 	\textbf{45.35} &	\textbf{64.25} &	\textbf{53.43} &	\textbf{20.19} \\
                    \quad NJUST KMG & SWIN-L  
                    & \makecell{Objects365V2, OpenImageV6, GoldG, V3Det, COCO2014, COCO2017, \\ LVISV1, GRIT, RefCOCO, RefCOCO+, RefCOCOg, gRef-COCO} & 32.56 	&   50.21 	&   34.87 	&  15.16  \\
                    \quad zjyd\_cxy\_vision & SWIN-L  
                    & \makecell{Objects365V2, COCO2017, LVIS, GoldG, VG, OpenImagesV6, V3Det, 
                    \\  PhraseCut, RefCOCO, RefCOCO+, RefCOCOg, gRef-COCO} 
                    & 31.57 &	46.59 &	33.32 &	17.03\\
                    \bottomrule
                    
                \end{tabular}
            }
            \vspace{-2mm}
            \label{tab:all_res_10_shots}
        \end{table}
        
\subsection{Empirical Analysis of Results}
We evaluate several popular VLMs on the nuImages Foundational FSOD (10-shot) benchmark and present salient insights from Table \ref{tab:all_res_10_shots} below. 

{\bf Zero-Shot Detection.} Somewhat unsurprisingly, we find that (1) greater pre-training data scale and diversity, along with (2) larger backbones result in better zero-shot performance. Notably, GLIP achieves 17.01\% zero-shot performance, surpassing all other methods trained with less data and smaller backbones. 

{\bf Prompt Engineering.} We attempt to improve zero-shot performance using synonyms for class names derived from the annotator textual instructions. We see minor improvements (e.g., Detic improves from $14.40$ mAP $\rightarrow 14.92$ mAP), indicating that leveraging rich textual descriptions beyond class names can improve concept alignment.

{\bf Federated Fine-Tuning.} Standard fine-tuning is sub-optimal for FSOD, as all unannotated classes are treated as negatives. 
Therefore we use our zero-shot predictions to generate pseudo-labels on training images. We extract pseudo-negatives based on these pseudo-labels by identifying classes {\em not} in each image (by using detector confidence scores), and leverage pseudo-negatives in our fine-tuning. Notably, we improve over Detic's standard fine-tuning by $1.15$ mAP ($16.09$ mAP $\rightarrow 17.24$ mAP). 

{\bf Multi-Modal Prompting.} We observe that Multi-Modal Prompting ({\tt MQ-GLIP}) achieves the best performance ($21.42$ mAP) out of all open-source methods in Table \ref{tab:all_res_10_shots}. We attribute this to its large pre-trained dataset, bigger backbone (SWIN-L) and multi-modal prompts used during inference. Notably, the benefit of multi-modal prompts can be seen by comparing {\tt MQ-GLIP} ($21.42$ mAP) against {\tt MQ-GLIP-Image} ($14.07$ mAP), which uses visual prompting and {\tt MQ-GLIP-Text} ($17.01$ mAP), which uses language prompting. Interestingly, {\tt MQ-GLIP} does not require gradient-based fine-tuning, which differs from all existing conventional few-shot methods. Therefore, we posit that future few-shot methods should further explore in-context learning. Just as multi-modal examples aid human annotator alignment, multi-modal prompting significantly improves VLM concept alignment.


\begin{figure}[t]
    \centering
    \includegraphics[width=1.0\linewidth]{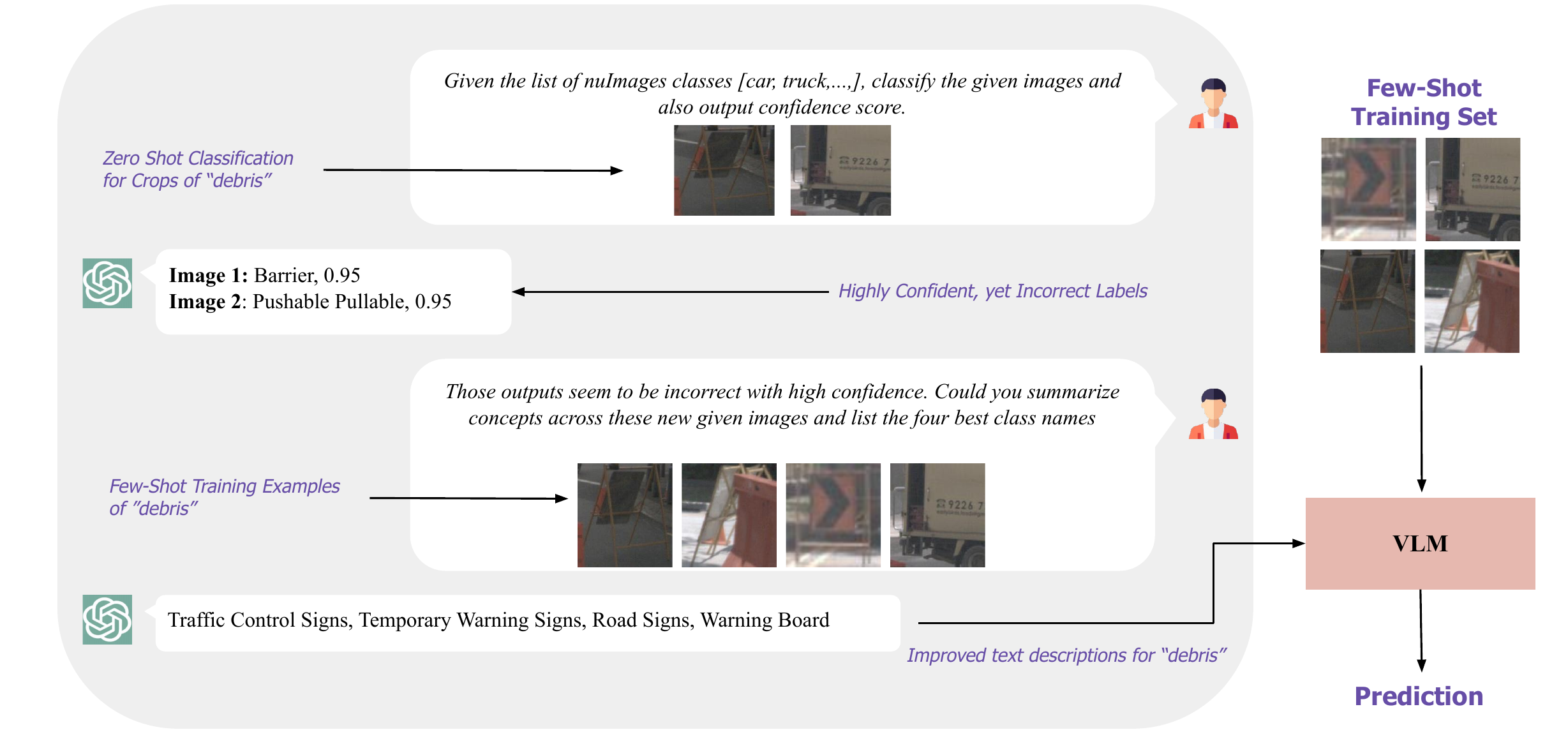}
    \vspace{-3mm}
    \caption{\small
    \textbf{Iteratively Prompting ChatGPT.} Despite its large-scale pre-training, multi-modal models like ChatGPT-4o also suffers from concept misalignment. Specifically, GPT-4o makes highly confident but incorrect predictions for {\tt debris}. We propose an iterative prompting strategy to better align the model to a target concept. Given a few visual examples per-class from the training-set, we query ChatGPT to use its ``web-scale knowledge'' to generate text descriptions. We then augment the input to MQDet to incorporate this additional context for zero-shot evaluation.}
    \label{fig:gpt_refine}
\end{figure}

{\bf Multi-Modal Chat Agents}. As shown in Figure~\ref{fig:gpt_refine},  we explore the idea of iteratively prompting multi-modal chat assistants like ChatGPT to mimic the real-world workflow of human annotators.  Given the strong performance of GPT-4o for general visual understanding, we repurpose it for our task by prompting the model to re-classify image patches from Detic's RPN. Specifically, we ask GPT-4o to predict a class and confidence for each image crop. Surprisingly, we observe reasonable performance (9.95 mAP). However, we find that GPT-4o often incorrectly classifies many image crops with high confidence. Therefore, we prompt GPT-4o to generate its own text descriptions of the few-shot examples according to its ``web-scale knowledge''. Finally, we use the class names, generated text descriptions, and few-shot visual examples to prompt MQDet to predict new instances of target classes in the test-set. We find that expanding MQDet's in-context prompt with class names, ChatGPT generated text descriptions, and few-shot visual examples improves performance by 0.67\% ($21.42$ mAP $\rightarrow$ $22.09$ mAP) over the MQ-GLIP baseline. Interestingly, although {\tt debris} accuracy does not change when prompted with generated text descriptions, {\tt pushable pullable} ($3.6$ AP $\rightarrow$ $15.29$ AP) and {\tt barrier} ($11.6$ AP $\rightarrow$ $15.31$ AP) accuracy improve significantly. We posit that this improvement is due to the reduction in confusion (or over-confident incorrect predictions) between {\tt debris} and {\tt pushable-pullable} (and {\tt barrier}). Surprisingly, a top submission to our CVPR challenge also used ChatGPT to generate meaningful text descriptions to improve {\em concept alignment}.

{\bf CVPR 2024 Challenge}. Our inaugural Foundational FSOD competition (hosted on \href{https://eval.ai/web/challenges/challenge-page/2270/evaluation}{Eval AI}) received submissions from seven teams (some submissions are private). We present a ranked list of participants at the close of our competition on June 7th 2024 AOE in Table \ref{tab:competition}. Notably, three teams beat our baselines, with the winning team achieving $45.35$ AP! Unfortunately, the top performing team was not willing to publicly share details
about their method. We summarize contributions from the other two top teams below.

\begin{table}[b]
            \vspace{-2mm}
            \centering
            \caption{\small
            \textbf{CVPR 2024 Foundational FSOD Competition Results}.
            }
            \vspace{-2mm}
            \def\arraystretch{1.0}%
            \resizebox{0.85\linewidth}{!}{
            
                \begin{tabular}{@{}l@{\ \ \ \ \ \ \ \ \ \ \ \ \ \ \ \ \ \ \ \ }c@{\ \ \ \ \ \ \ \ \ \ \ \ \ \ \ \ \ }c@{\ \ \ \ \ \ \ \ \ \ \ \ \ \ \ \ \ }c@{\ \ \ \ \ \ \ \ \ \ \ \ \ \ }c@{}}
                    \toprule
                    \multirow{2}{*}{Team Name} & \multicolumn{4}{c}{Average Precision (AP)}  \\
                    \cmidrule(l){2-5} 
                                        & {\tt All} & {\tt Many}    & {\tt Medium}   & {\tt Few}    \\ 
                    

                    \midrule
                     PHP\_hhh & 45.35  	&   64.25 	&   53.43 	&  20.19  \\
                     NJUST KMG &  32.56 	&   50.21 	&   34.87 	&  15.16  \\
                     zjyd\_cxy\_vision &  31.57 	&  46.59  	&   33.32 	&  17.03  \\
                     Baseline (MQ-GLIP) & 21.51  	&  32.25  	&   23.35 	&  10.41  \\
                     team\_anon &  17.36 	&  25.29  	&   21.93 	&  5.42  \\
                     youyouqiu &  13.16 	&  11.29  	&  19.20  	&  7.68  \\
                     zhao &  11.38 	&  11.16  	&   16.76 	&  5.30  \\
                     zjdcxy & 7.80  	&  5.44  	&   13.43 	&  3.20  \\
                    \bottomrule
                \end{tabular}
            }
            \vspace{-4mm}
            \label{tab:competition}
        \end{table}

{\bf NJUST KMG} presents a method that leverages both VLMs and multi-modal chat agents for Foundational FSOD. To address the challenge of misalignment between GroundingDINO and the target concepts of interest, authors generate descriptive referential expressions by prompting ChatGPT to provide descriptive terms for each few-shot instance. The best referential expression for each category is selected by maximizing the Intersection over Union (IoU) between predictions and the ground truth in the few-shot training set. These referential expressions are then used to generate pseudo-labels for all training images. Lastly, GroundingDINO is fine-tuned on a combination of pseudo-labels and ground-truth instances. The full technical report is available \href{https://www.neeharperi.com/files/njustkmg_techreport_cvprw24.pdf}{here}.

{\bf ZJYD CXY Vision} proposes Instruction DINO (ISD), a DETR-based detector architecture that incorporates early fusion of image and text information using a Swin-L visual backbone and EVA02-CLIP-L text encoder. Authors use VLMs like CLIP, TAP, and Llava for negative sample generation (similar to our Federated Fine-Tuning). Authors find that prompt tuning and text encoder fine-tuning generalize better than visual encoder fine-tuning. Similar to NJUST KMG, authors first generate pseudo-label annotations for unlabeled categories before fine-tuning on a combination of pseudo-labels and ground truth instances. The final method combines prompt tuning and negative sampling, significantly improving mAP. The full technical report is available \href{https://www.neeharperi.com/files/zjydcxyvision_techreport_cvprw24.pdf}{here}.

\begin{figure*}[b]
    \centering
    \includegraphics[width=1.0\linewidth]{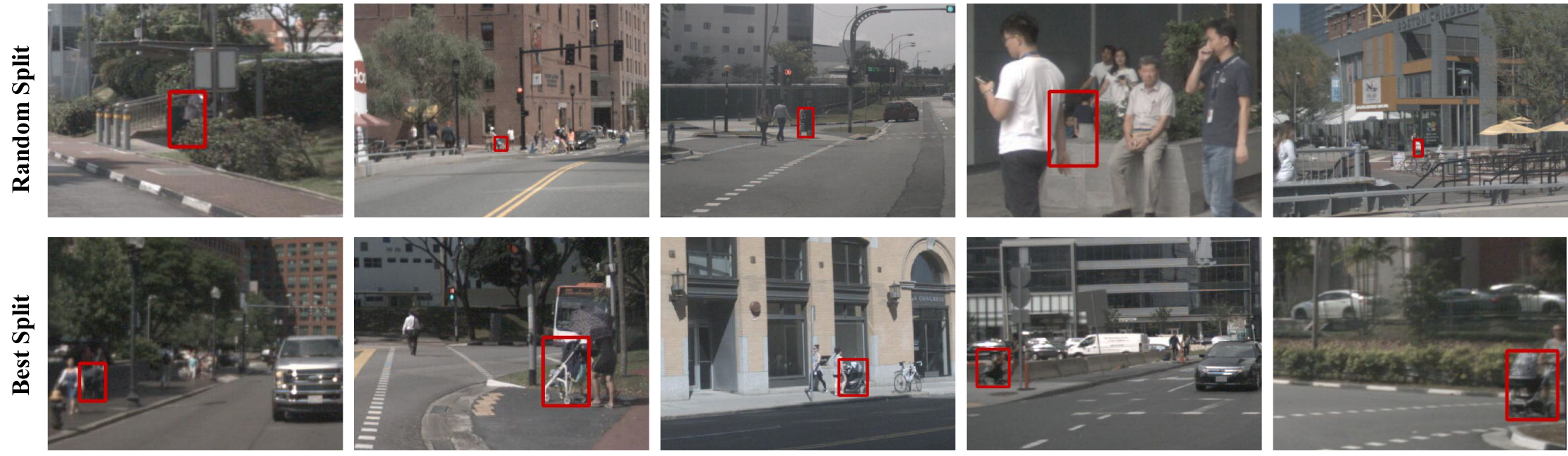}
    \vspace{-4mm}
    \caption{\small
    \textbf{Visualizing Random and Best Split}. In the top row, we visualize the 5-shot training examples of \textbf{strollers} from a \textit{random split}. Similarly, we visualize the 5-shot training examples from the \textit{best split} in the bottom row. We observe that strollers in the \textit{random split} are often occluded, small in size and blurry, making few-shot learning harder. On the other hand, the \textit{best split} examples are larger, have better visual quality and are relatively un-occluded. This visual difference directly translates into better few-shot performance. We achieve $\mathbf{13.09}$ \textbf{Stroller AP} for the \textit{random split} and $\mathbf{18.54}$ \textbf{Stroller AP} for the \textit{best split}. We show a more comprehensive evaluation in Table~\ref{tab:best_split}.
    }
    \label{fig:stroller_best_split}
    \end{figure*}

\begin{table}[t]
            \centering
            \caption{\small
            \textbf{Random Split vs ``Best'' Split}. We construct the ``best" split by selecting per-class few-shot examples that lead to the highest performance on a held-out set. Unsurprisingly, the best split performs better than any random split, especially for very limited data settings (e.g., $5$-shot detection). This evaluation setting closely mimics how human annotators are ``aligned'' to target concepts, since annotator guides are constructed using hand-picked iconic visual examples.
            }
            \vspace{-2mm}
            \def\arraystretch{1.0}%
            \resizebox{0.99\linewidth}{!}{
            
                \begin{tabular}{@{}l@{\ \ \ \ \ \ \ \ \ \ \ \ \ \ \ \ \ \ \ \ }c@{\ \ \ \ \ \ \ \ \ \ \ \ \ \ \ \ \ }c@{\ \ \ \ \ \ \ \ \ \ \ \ \ \ \ \ \ }c@{\ \ \ \ \ \ \ \ \ \ \ \ \ \ }c@{}}
                    \toprule
                    \multirow{2}{*}{Approach} & \multicolumn{4}{c}{Average Precision (AP)}  \\
                    \cmidrule(l){2-5} 
                                        & {\tt All} & {\tt Many}    & {\tt Medium}   & {\tt Few}    \\ 
                    

                    \midrule
                        Detic (Zero-Shot) \cite{zhou2022detecting} &  14.40	&25.83	&16.59	&2.32    \\
                    \midrule
                        Detic w/ Federated Fine-Tuning \textit{(5-shots, Random Split)} & 16.58	&27.12	&19.71	&4.13 \\
                        Detic w/ Federated Fine-Tuning \textit{(5-shots, Best Split)} & \textbf{18.30}	& \textbf{28.66}	&\textbf{21.81}	&\textbf{5.56}\\
                    \midrule
                        Detic w/ Federated Fine-Tuning \textit{(10-shots, Random Split)} & 17.24	&28.07	&20.71	&4.18  \\
                        Detic w/ Federated Fine-Tuning \textit{(10-shots, Best Split)} &\textbf{18.24}	&\textbf{28.63}	&\textbf{22.00}	&\textbf{5.19} \\
                    \midrule
                        Detic w/ Federated Fine-Tuning \textit{(30-shots, Random Split)} &18.64	&\textbf{29.13}	&22.44	&5.46	 \\
                        Detic w/ Federated Fine-Tuning \textit{(30-shots, Best Split)} &\textbf{18.75}	&28.07	&\textbf{23.18}	&\textbf{5.81}	 \\ 
                    \bottomrule
                \end{tabular}
            }
            \vspace{-4mm}
            \label{tab:best_split}
        \end{table}
        
\subsection{Analysis of Iconic Few-Shot Images}\label{sec:best_split}

The specific examples used during few-shot fine-tuning significantly impacts target class performance \cite{wang2020frustratingly}. However, prior work constructs few-shot splits by randomly sampling $K$ examples per class. In contrast, when creating annotator {\em instructions}, selecting the right examples to ``align" human annotators \cite{chang2023thinking} to subtle aspects of the target concept is carefully considered. To more closely match VLM {\em concept alignment} with human annotator alignment, we design a simple algorithm to construct the best $K$-shot split for fine-tuning. This allows us to understand which examples are most informative and measure an upper bound in performance.

We construct our \textit{best split} by picking examples corresponding to the best class-wise performance, based on the evaluation of each split on a held-out validation set. For instance, out of $3$ random splits for the $5$-shot task, one may pick {\tt car} examples from split $1$, {\tt bicycle} from split $3$ and {\tt debris} from split $2$. In Table~\ref{tab:best_split}, we observe that the \textit{best-split} performance is always better than its random counterpart. As expected, the choice of examples in $5$-shot case is more important than the $30$-shot case ($1.72$ AP difference for $5$-shot vs $0.11$ AP for $30$-shots). We visualize the difference in the splits for {\tt stroller}s in nuImages (cf. Figure~\ref{fig:stroller_best_split}). Unsurprisingly, iconic examples are large and unoccluded.

\subsection{Limitations and Future Work} \label{sec:limitations}
Despite using VLMs pre-trained on large-scale datasets, we find that performance for rare categories (defined by the cardinality of each class in the original dataset) is considerably lower than for common classes. We posit that VLMs are pre-trained with imbalanced data which includes many examples of common categories like {\tt truck} but few examples of rare categories like {\tt stroller}~\cite{parashar2024neglected}.
Our work does not explicitly improve detection performance on rare classes.
Interestingly, since VLMs like Detic \cite{zhou2022detecting}, GLIP \cite{li2021grounded}, and GroundingDINO \cite{liu2023grounding} are trained with different data sources, each model has dramatically different zero-shot performance on novel categories like {\tt stroller}. Ensembling predictions from different VLMs may yield better detection accuracy for rare categories. 
In addition, although our work motivates the use of rich textual descriptions found in instructions for multi-modal alignment, our current results use only nouns (class names and synonyms) as text prompts.

{\bf Benchmarking in the Era of Foundation Models}. Although we argue that pre-training on large-scale data will be the key enabler for generalization to the open-world, understanding how to appropriately benchmark such methods remains challenging. It is readily accepted that in order to accurately evaluate generalization, one should not train on test data. However, it is difficult to guarantee that foundation models have never seen our specific test data. We address this in our challenge by explicitly prohibiting participants from training on nuImages (and nuScenes). 
However, should we allow participants to train on similar in-domain data (e.g., other autonomous vehicle datasets such as Argoverse~\cite{Argoverse2})? 
We argue `yes'! 
With enough scale, novel test examples may still be similar to the training set. 

{\bf Out-of-Domain Benchmarks}. Another way to address benchmarking is to collect test scenarios that are {\em designed} to be dissimilar from internet images. For example, out-of-domain images may include medical data (though foundational performance is still surprisingly effective~\cite{Wang2022MedCLIPCL}). We admittedly did not take this route, since urban imagery is similar to images found online and arguably many applications of interest fall under this category. Moreover, there exist ample opportunity for technical innovation in this setting (as suggested by our CVPR 2024 challenge results!). Alternatively, one can 
manually collect and sequester images that will never be released on the internet. Since ensuring privacy may itself be challenging, yet another approach is to leverage the continual learning paradigm \cite{lin2022continual}, where new test sets are continually constructed over time. 

{\bf Comparing Models}. Fairly comparing foundation models requires careful consideration. Although accuracy is a valuable metric, it is intrinsically tied to the scale of pre-training data and model architecture. Notably, the LLM community already ranks models via a Pareto frontier of accuracy vs. parameter count. We advocate for a similar approach for Foundational FSOD that considers backbone architecture (e.g., ResNet-50 vs. Swin-B) and pre-training datasets (e.g., CC4M, GoldG, LVIS).

\section{Conclusion}
We revisit few-shot object detection (FSOD) with vision-language models (VLMs) and find that zero-shot inference from web-scale VLMs significantly outperforms leading FSOD methods. However, such foundational models do not fully address few shot recognition because of the {\em concept alignment} problem; particular concepts in target applications may be different than their use on web-scale datasets. Just as human annotators require concept alignment via multi-modal text and visual examples, we argue that VLMs should be aligned with such few-shot data, formalizing the problem of Foundational FSOD. 


\section{Acknowledgements}
This research was supported by Bosch Center for AI (BCAI). We thank BCAI for their financial support and resources, which made this work possible. This work was also supported in part by the Institute of Collaborative Innotation and the University of Macau (SRG2023-00044-FST) and the NSF GRFP (Grant No. DGE2140739).

\newpage 

\printbibliography

    

\appendix

\section{Baseline Implementation Details}
\begin{figure}[h]
    \centering
    \includegraphics[width=1.0\linewidth]{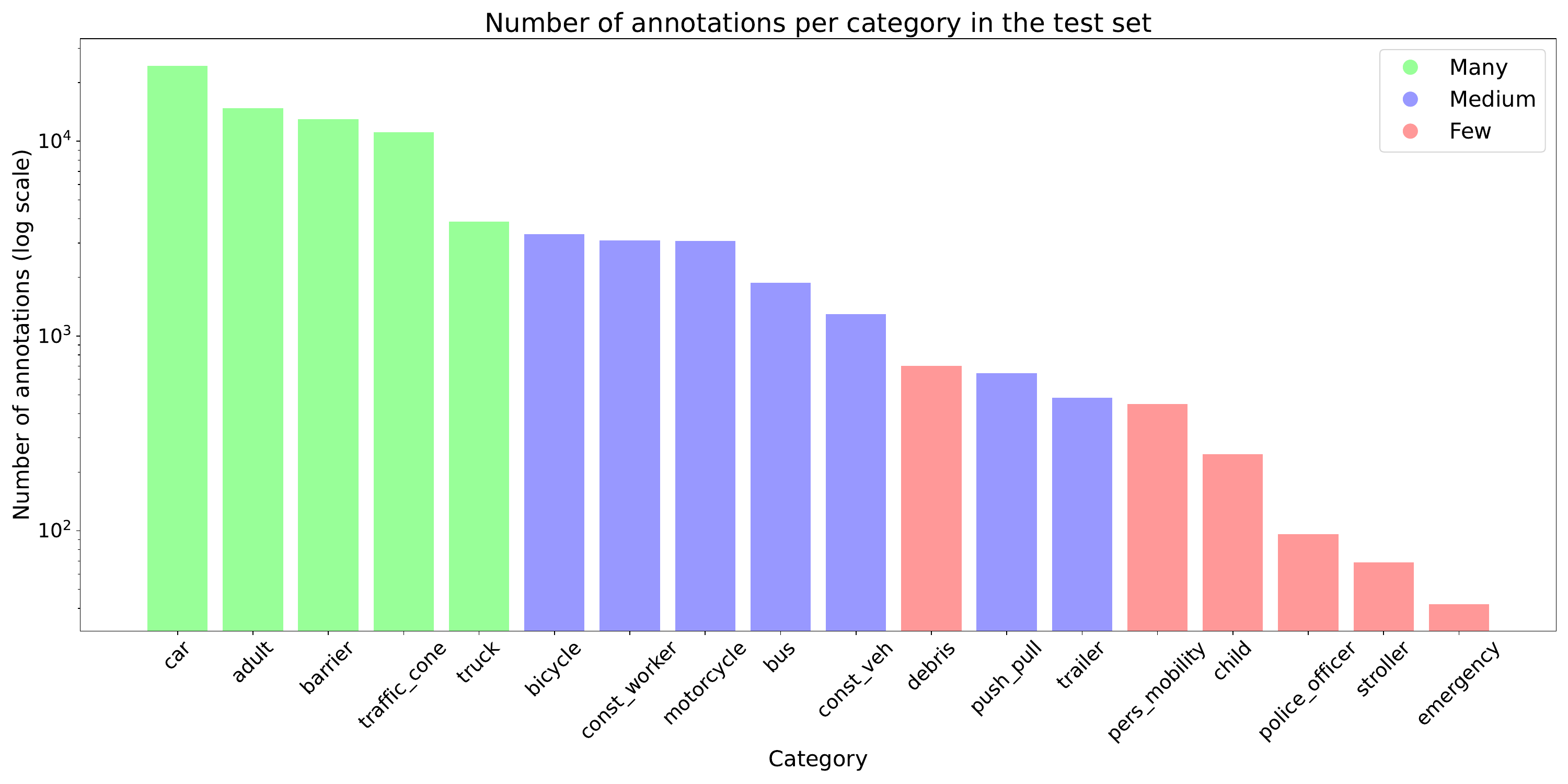}
    \vspace{-5mm}
    \caption{\small We visualize the distribution of classes in out test-set compared to the cardinalities of classes in the full nuImages val-set. Notably, our sub-sampling strategy of selecting validation images that have at least one annotation from {\tt medium} or {\tt few} classes does not significantly alter the true distribution.}
    \label{fig:test_set_viz}
\end{figure}

We repurpose nuImages (CC BY-NC-SA 4.0) for all few-shot experiments in the main paper. We evaluate detection performance using $1600 \times 900$ images across 18 classes for all models tested. We create three random splits for each of $K=\{5,10,30\}$-shots following the data creation process from \cite{wang2020frustratingly} and report results averaged across these three seeds. Our test-set is a subset of the (densely annotated) nuImages val-set. We construct our test-set to only include validation images which have at least one annotation from the {\tt Few} or {\tt Medium} cohorts (cf. Fig~\ref{fig:test_set_viz}). We train all baselines with one RTX 3090 GPU. Our baseline code is available on \href{https://github.com/anishmadan23/foundational_fsod}{GitHub} and dataset splits are available on \href{https://huggingface.co/anishmadan23/foundational_fsod/tree/main}{HuggingFace}.\\

\textbf{Prompt Engineering}: We leverage rich text descriptions provided by the annotator instructions to select synonyms for each nuImages class. We manually identify the best performing synonyms in Table \ref{tab:cls_syn}. At test time, we compute the average text embedding of all synonyms to improve classification accuracy. 

 \begin{table}[t]
            \centering
            \caption{\small
            \textbf{Synonyms used for Prompt Engineering.} We manually inspect the nuImages annotator instructions to derive a set of synonyms to improve classification accuracy.}
            \vspace{-2mm}
            \resizebox{0.9\linewidth}{!}{
                \begin{tabular}{@{}l@{\ \ \ \ \ \ \ \ \ \ \ \ \ \ \ \ \ \ \ \ \ \ \ }l@{}}

                    \midrule
                    \textbf{Original Classes} & \textbf{Class Names with Synonyms} \\
                    \midrule
                    {\tt car} & {\tt car} \\
                    \midrule
                    {\tt truck} & {\tt truck}, {\tt pick-up}, {\tt lorry}, {\tt semi-tractor} \\
                    \midrule
                    {\tt construction\_vehicle} & {\tt construction\_vehicle}, {\tt crane} \\
                    \midrule
                    {\tt bus} & {\tt bus}, {\tt bendy\_bus}, {\tt rigid\_bus} \\
                    \midrule
                    {\tt trailer} & {\tt trailer} \\
                    \midrule
                    {\tt emergency} & {\tt emergency}, {\tt ambulance}, {\tt police\_car}, {\tt police\_motorcycle} \\
                    \midrule
                    {\tt motorcycle} & {\tt motorcycle} \\
                    \midrule
                    {\tt bicycle} & {\tt bicycle} \\
                    \midrule
                    {\tt adult} & {\tt adult}, {\tt person} \\
                    \midrule
                    {\tt child} & {\tt child} \\
                    \midrule
                    {\tt police\_officer} & {\tt police\_officer} \\
                    \midrule
                    {\tt construction\_worker} & {\tt construction\_worker} \\
                    \midrule
                    {\tt personal\_mobility} & {\tt personal\_mobility}, {\tt skateboard}, {\tt segway}, {\tt scooter} \\
                    \midrule
                    {\tt stroller} & {\tt stroller} \\
                    \midrule
                    {\tt pushable\_pullable} & {\tt pushable\_pullable}, {\tt wheel\_barrow}, {\tt garbage\_bin}, {\tt cart} \\
                    \midrule
                    {\tt barrier} & {\tt barrier}, {\tt K-rail}, {\tt fence}, {\tt bollard}, {\tt guard\_rail} \\
                    \midrule
                    {\tt traffic\_cone} & {\tt traffic\_cone} \\
                    \midrule
                    {\tt debris} & {\tt debris}, {\tt trash\_bag} \\
                    \bottomrule
                \end{tabular}
                }
        \label{tab:cls_syn}
        \vspace{-3mm}
    \end{table}

\textbf{Language Prompt Tuning} We train GLIP (SWIN-L backbone) for our prompt tuning experiments for $60$ epochs with a learning rate of $0.025$, batch size of $4$, and weight decay of $0.25$.

\textbf{Federated Fine-tuning}.
We use Detic (Swin-B backbone) pre-trained on {\tt LVIS + COCO} and ImageNet-21k data for our federated fine-tuning experiments (described in detail in the next section). We use a batch size of $8$ and an AdamW optimizer with learning rate of $3.75e-6$. We fine-tune this model for $8000$ iterations on nuImages. We sample $6$ categories for each training image, i.e $|S|=6$ for the FedLoss and InvFedLoss experiments. We derive negatives from pseudolabels with atleast $20\%$ confidence for the Psuedo-Negative experiment.

\textbf{Multi-Modal Prompting}. We use MQDet ({\tt text-only, vision-only, text + vision} for our in-context learning baselines. Unlike the original code base, we tokenize our few shot examples instead of using random queries. Note that zero-shot results for MQ-GLIP-Text and GLIP-L are the same since these models are identical. 


\section{Analysis of Federated Fine-Tuning}
Prior works follow the $K$-shot dataset creation process established by \cite{wang2020frustratingly}. Importantly, each image in the dataset is exhaustively annotated for a subset of all classes. Recall, a federated dataset is also comprised of images that are exhaustively annotated for a specific category. This suggests that we can leverage existing insights about federated datasets \cite{gupta2019lvis, zhou2021probablistic} to train better few-shot object detectors. 

\textbf{Fine-Tuning with FedLoss}. We fine-tune Detic with Federated Loss (FedLoss)  \cite{zhou2021probablistic} using a subset $S$ of classes $C$ for each training image. Specifically, we use a binary cross-entropy loss on all classes in $S$ and ignore classes outside of $S$ during training. $S$ is comprised of the ground-truth annotation class along with randomly sampled negative classes for each image. We sample these negative classes in proportion to their square-root frequency in the training set. We find that probablistically sampling negatives rather than labeling all unannotated classes as negatives improves fine-tuning results, reliably beating zero-shot performance. Importantly, although FedLoss has been explored in the context of long-tailed detection, applying it to FSOD provides considerable performance improvements, reaffirming that FSOD benchmarks are actually federated datasets.

 \textbf{Fine-Tuning with Pseudo-Negative Federated Loss (Ours).} Despite the effectiveness of FedLoss, probablistically sampling negatives using dataset-wide statistics is sub-optimal because it does not consider the content of each image. We can improve the accuracy of sampled negatives with pseudo-labels to determine which classes are likely \emph{not} in a particular image. If the maximal score for any class prediction is less than a threshold, we consider this class to be a negative. Using zero-shot model predictions to identify pseudo-negatives yields better results than simply using dataset-wide statistics. We find that this strategy works the best. We present pseudo-code in Alg. \ref{alg:psuedoneg-fedloss}. All federated fine-tuning results in the main paper are trained with psuedo-negative federated loss.

\RestyleAlgo{ruled}
    \SetKwComment{Comment}{/* }{ */}
    \begin{algorithm}[htb!]
    \caption{\small
    Psuedo-Negative Federated Loss
 }\label{alg:psuedoneg-fedloss}
    \begin{pythonic}
# Inputs
# img: Randomly Sampled Image
# all_classes: All Classes in Dataset
# gt: Ground Truth Annotations for img
# gt_classes: List of Classes in gt
#
# Outputs
# loss: Psuedo-Negative Federated Loss
#
# Functions
# filter: Returns All Predictions w/ 
#        Confidence > Threshold 
# get_neg: Returns List of Classes Not 
#         In Pseudo-Positives
# or: Set Union Operation
# BCE: Binary Cross Entropy Loss

#Step 1: Compute Predictions
#        and Filter by Confidence
pred = Detector(img) # predictions
pseudo_pos = filter(pred, thresh=0.2)

#Step 2: Get Pseudo-Negatives for Image
neg_classes = get_neg(pseudo_pos, all_classes)
select_classes = or(neg_classes, gt_classes)

#Step 3: Compute Deterministic Federated Loss 
#         w/ Pseudo-Negatives
loss = 0
for cls in select_classes:    
    pred_cls = pred[cls] #predictions for cls
    gt_cls = gt[cls] #ground-truth for cls
    loss += BCE(pred_cls, gt_cls)
return loss
    \end{pythonic}
    \end{algorithm}

\begin{table}[t]
            \centering
            \caption{\small
            \textbf{Analysis of nuImages Upper Bound Performance}. 
            We compare the accuracy of our proposed approach against upper bounds computed for the FSOD task. Our pseudo-negatives strategy approaches the performance of using ground-truth negatives, demonstrating that pesudo-labels can provide a reliable signal about negatives, especially across classes with {\tt many} and {\tt medium} examples. The performance gap between our best method and exhaustive annotations can be attributed to the large number of additional annotations, particularly for classes with {\tt many} and {\tt medium} examples. Compared to the baseline (14.3 AP), our approach (16.7 AP) closes the gap to the (18.5 AP) upper-bound by over 50\%.
            }
            \vspace{-2mm}
            \def\arraystretch{1.0}%
            \resizebox{0.99\linewidth}{!}{
            
                \begin{tabular}{@{}l@{\ \ \ \ \ \ \ \ \ \ \ \ \ \ \ \ \ \ \ \ }c@{\ \ \ \ \ \ \ \ \ \ \ \ \ \ \ \ \ }c@{\ \ \ \ \ \ \ \ \ \ \ \ \ \ \ \ \ }c@{\ \ \ \ \ \ \ \ \ \ \ \ \ \ }c@{}}
                    \toprule
                    \multirow{2}{*}{Approach} & \multicolumn{4}{c}{\textbf{10 Shots}: Average Precision (AP)}  \\
                                        & {\tt All} & {\tt Many}    & {\tt Medium}   & {\tt Few}    \\ 
                    

                    \midrule
                        Detic (Zero-Shot) \cite{zhou2022detecting} & 14.26 & 27.28 & 15.15 & 2.36  \\
                    \midrule 
                    
                        \quad + Standard Fine-Tuning &  15.53  & 26.01 & 18.02 &	3.88  \\
                        \quad w/ FedLoss &  15.57 & 27.20 &	18.13 &	2.89	   \\
                        \quad w/ Pseudo-Negatives & \textbf{16.67} & \textbf{29.15} & \textbf{18.71}	& \textbf{3.90}  \\
                    \midrule
                         \quad w/ True Negatives ({\em Oracle}) & 16.99 & 29.60 & 18.94 & 4.21  \\
                        \quad w/ Exhaustive Annotations ({\em Oracle}) &  18.51 & 33.51 &	20.30 &	3.93	 \\
                     
                    \bottomrule
                \end{tabular}
            }
            \vspace{-3mm}
            \label{tab:nuimages_10_shots}
        \end{table}
        
\textbf{Oracle Performance Analysis}. We empirically validate the effectiveness of our pseudo-negative federated loss by computing the upper bound performance when given access to ground-truth negatives and exhaustive annotations for the few-shot data split. Recall, nuImages is exhaustively annotated, but is repurposed for Foundational FSOD. 

To compute the set of ground-truth negatives for each image, we use exhaustive ground-truth annotations to determine which categories are not present. Training with ground-truth negatives provides an upper bound on our pseudo-negatives experiment. 
Next, we train using exhaustive ground-truth annotations to provide an upper bound for the specific set of images used during training. In addition, this experiment highlights the performance gap between having exhaustive negatives and exhaustive annotations.

Table~\ref{tab:nuimages_10_shots} shows that using pseudo-negatives nearly matches the true negative upper bound (16.67 AP vs 16.99 AP). This demonstrates that we are able to reliably estimate negatives in an image, alleviating the problem of learning with sparse annotations. Training with exhaustive annotations yields significantly better results for  {\tt many} and {\tt medium} classes. This is unsurprising because the 10-shot FSOD benchmark includes 10 car annotations, while the exhaustively annotated set includes over 550 car annotations! 

Despite strong performance on classes with {\tt many} and {\tt medium}, the upper bound for classes with {\tt few} examples remains low (4.21 AP and 3.93 AP). 
Given the success of training with pseudo-negatives, a natural next-step is to train with pseudo-positives. Our preliminary results suggest that incorporating pseudo-positives does not provide significant improvement over simply training with pseudo-negatives. We posit that training with incorrect pseudo-positives may incur a higher penalty than training with incorrect pseudo-negatives. This is a promising direction for future work.

\section{Impact of Box-Level Supervision for Foundational FSOD}
We evaluate the importance of using bounding-box supervised data in pre-training. Unlike Detic, which trains on box-supervised data from {\tt LVIS, COCO} and image-text data from {\tt ImageNet21k}, RegionCLIP\cite{zhong2022regionclip} only pre-trains on image-text pairs from the Conceptual Captions (CC3M) dataset \cite{cc4m}. We report RegionCLIP's zero-shot and fine-tuning performance on nuImages averaged over $3$ random splits in Table~\ref{tab:regionclip1}. Detic zero-shot outperforms RegionCLIP zero-shot by $\sim12$ AP ($14.26$ vs $2.34$). While fine-tuning RegionCLIP improves overall performance, Detic achieves higher accuracy for $K=\{5,10,30\}$ shots. This highlights the importance of supervision type (e.g. box-supervised data) and data scale used for pre-training. 

Next, we conduct further analysis to diagnose why RegionCLIP zero-shot inference performs so poorly on nuImages (Table~\ref{tab:regionclip_diagnostic}). RegionCLIP relies on an RPN pre-trained on box-supervised data like {\tt LVIS-base} to extract regions for pre-training. Notably, RegionCLIP (w/ {\tt LVIS-RPN}: $2.34$ AP) suffers from poor foreground-vs-background classification compared to Detic. We validate this hypothesis by evaluating RegionCLIP (w/ {\tt GT-RPN)} to measure classification performance. Surprisingly, RegionCLIP achieves significantly higher accuracy ($26.44$ AP), confirming that RegionCLIP struggles to distinguish between foreground and background in nuImages. This observation highlights the challenge of working with nuImages categories, further motivating our Foundational FSOD benchmark. 

Lastly, we evaluate RegionCLIP's performance with {\tt Detic-RPN}. Notably, we observe that the performance improves over RegionCLIP w/ {\tt LVIS-RPN} demonstrating that reducing the number of false positive proposals yields better performance. Furthermore, we filter out low confidence Detic proposals , i.e $<0.5$ objectness score (w/ {\tt Detic-RPN, 0.5}) and find that this doubles RegionCLIP's zero-shot performance to $7.64$ AP.

\begin{table}[b]
            \centering
            \caption{\small
            \textbf{RegionCLIP Experiments}. RegionCLIP zero-shot inference performs much worse than Detic. While fine-tuning improves RegionCLIP's performance, it still lags far behind Detic. We posit that this performance difference can be attributed to Detic's box-supervised pre-training and use of language cues from CLIP embeddings.
            }
            \vspace{-3mm}
            \def\arraystretch{1.0}%
            \resizebox{0.99\linewidth}{!}{
            
                \begin{tabular}{@{}l@{\ \ \ \ \ \ \ \ \ \ \ \ \ \ \ \ \ \ \ \ }c@{\ \ \ \ \ \ \ \ \ \ \ \ \ \ \ \ \ }c@{\ \ \ \ \ \ \ \ \ \ \ \ \ \ \ \ \ }c@{\ \ \ \ \ \ \ \ \ \ \ \ \ \ }c@{}}
                    \toprule
                    \multirow{2}{*}{Approach} & \multicolumn{4}{c}{Average Precision (AP)}  \\
                                        & {\tt All} & {\tt Many}    & {\tt Medium}   & {\tt Few}    \\ 
                    

                    \midrule
                        RegionCLIP (\textit{Zero-Shot}) \cite{zhong2022regionclip} & 2.34 & 3.33 &	3.45 & 0.22  \\
                        Detic (\textit{Zero-Shot}) \cite{zhou2022detecting} & \textbf{14.26} & \textbf{27.28} & \textbf{15.15} & \textbf{2.36}  \\
                    \midrule
                         RegionCLIP (\textit{Fine-Tuning, 5 shots}) \cite{zhong2022regionclip}  & 3.61 & 6.20 & 4.63 &  0.26 \\
                        Detic (\textit{Fine-Tuning, 5 shots}) \cite{zhou2022detecting} & \textbf{14.50} &	\textbf{24.09}	& \textbf{16.90} & 	\textbf{3.70}	\\
            
                    \midrule
                        RegionCLIP (\textit{Fine-Tuning, 10 shots}) \cite{zhong2022regionclip}  & 3.58 & 6.10 & 4.65 & 0.24	 \\
                        Detic (\textit{Fine-Tuning, 10 shots}) \cite{zhou2022detecting} & \textbf{15.28}	& \textbf{26.93} &	\textbf{18.00} &	\textbf{3.27}	 \\
            
                    \midrule
                        RegionCLIP (\textit{Fine-Tuning, 30 shots}) \cite{zhong2022regionclip}  & 3.57 & 6.13 & 4.61 & 0.22	 \\
                        Detic (\textit{Fine-Tuning, 30 shots}) \cite{zhou2022detecting} & \textbf{16.65}	& \textbf{27.45} &	\textbf{19.46} & 	\textbf{4.02}  \\
            
                    \bottomrule
                \end{tabular}
            }
            \label{tab:regionclip1}
        \end{table}

\begin{table}[t]
            \centering
            \caption{\small
            \textbf{Diagnosing RegionCLIP's Poor Zero-Shot Performance}. 
             RegionCLIP's zero-shot performance lags far behind Detic. Using RegionCLIP's classifier on ground-truth region proposals yields high performance, suggesting that RegionCLIP struggles to accurately distinguish between foreground-vs-background.}
            \vspace{-2mm}
            \def\arraystretch{1.0}%
            \resizebox{0.9\linewidth}{!}{
            
                \begin{tabular}{@{}l@{\ \ \ \ \ \ \ \ \ \ \ \ \ \ \ \ \ \ \ \ }c@{\ \ \ \ \ \ \ \ \ \ \ \ \ \ \ \ \ }c@{\ \ \ \ \ \ \ \ \ \ \ \ \ \ \ \ \ }c@{\ \ \ \ \ \ \ \ \ \ \ \ \ \ }c@{}}
                    \toprule
                    \multirow{2}{*}{Approach} & \multicolumn{4}{c}{Average Precision (AP)}  \\
                                        & {\tt All} & {\tt Many}    & {\tt Medium}   & {\tt Few}    \\ 
                    

                    \midrule
                        Detic \textit{(Zero-Shot)} \cite{zhou2022detecting} & 14.26 & 27.28 & 15.15 & 2.36  \\
                        GroundingDINO (Zero-Shot) \cite{liu2023grounding} & 11.44 & 17.42 & 14.08 & 3.38 \\
                        RegionCLIP \textit{(Zero-Shot)} w/ {\tt LVIS-RPN} \cite{zhong2022regionclip} & 2.34 & 3.33 &	3.45 & 0.22  \\
                    \midrule
                        RegionCLIP \textit{(Zero-Shot)} w/ {\tt Detic-RPN} \cite{zhong2022regionclip} & 3.79 &	6.68 &	4.01 & 	1.12 \\
                        RegionCLIP \textit{(Zero-Shot)} w/ {\tt Detic-RPN, 0.5} \cite{zhong2022regionclip} & 7.64 & 12.81	& 8.88 &	1.88 \\
                        RegionCLIP \textit{(Zero-Shot)} w/ {\tt GT-RPN} \cite{zhong2022regionclip} & 26.44	 & 45.33 &	32.25 &	3.92  \\

                    \bottomrule
                \end{tabular}
            }
            \label{tab:regionclip_diagnostic}
        \end{table}

\section{NuImages Annotator Instructions}\label{sec:nuim_ann_inst}
We present an example of the nuImages annotator instructions below. Notably, such annotator instructons are naturally few-shot (e.g. providing a few visual and textual examples describing the target concept), multi-modal, and contain both positive and negative examples. Our proposed Foundational FSOD benchmark, and pseudo-negative federated loss facilitate future work in leveraging rich annotator descriptions, allowing us to ``align'' VLMs much like how annotators must be ``aligned'' to subtle aspects of the target class. 
 
\begin{figure}[h]
    \centering
    \includegraphics[width=1.0\linewidth]{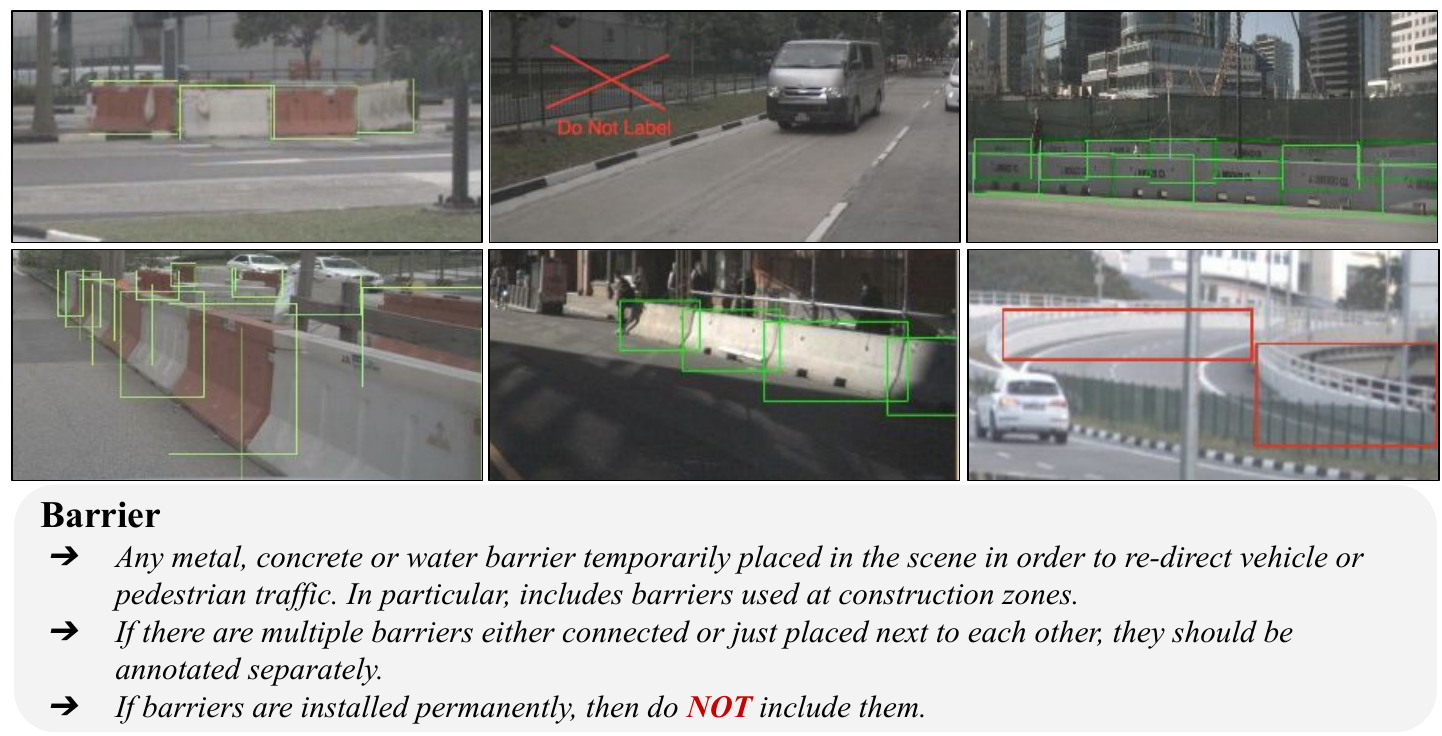}    
    \vspace{-4mm}
    \caption{\small \textbf{NuImages Annotator Instructions.} We include the \textbf{multi-modal} annotator instructions {\tt barrier}. Our proposed setup allows FSOD methods to learn such multi-modal examples, similar to how human annotators are taught the labeling policy. Importantly, annotators can also be provided with negative examples (in \purered{red}) for classes, i.e what \textbf{NOT} to label for a certain class. Crucially, our proposed fine-tuning with pseudo-negatives can easily accommodate such negative examples within the proposed setup.}
    \label{fig:nuscenes_anno_inst}
            \vspace{-2mm}
\end{figure}

\begin{table}[t]
            \centering
            \caption{\small
            \small {\bf Empirical Analysis of Baselines (5-shot) on nuImages}. 
            }
            \vspace{-2mm}
            \def\arraystretch{1.0}%
            \resizebox{0.99\linewidth}{!}{
            
                \begin{tabular}{@{}l@{\ \ \ \ \ \ \ \ \ \ \ \ \ \ \ }c@{\ \ \ \ \ \ \ \ \ }c@{\ \ \ \ \ \ \ \ \ \ \ \ \ }c@{\ \ \ \ \ \ \ \ \ }c@{\ \ \ \ \ \ \ \ \ }c@{\ \ \ \ \ \ \ \ \ }c@{}}
                    \toprule
                    \multirow{2}{*}{Approach} & \multirow{2}{*}{Backbone} & 
                    Pre-Train & \multicolumn{4}{c}{Average Precision (AP)}  \\
                      & & Data 
                                        & {\tt All} & {\tt Many}    & {\tt Med}   & {\tt Few}    \\ 
                   
                    \midrule
                        \textbf{Zero-Shot Detection} & & & & & & \\
                    \midrule
                        \quad RegionCLIP \cite{zhong2022regionclip}  & RN50 & CC3M & 2.50 &	3.20 &	3.80 &	0.40 \\
                        \quad Detic \cite{zhou2022detecting}  & SWIN-B & LVIS, COCO, IN-21K & 14.40	&25.83	&16.59	&2.32  \\
                        \quad GroundingDINO \cite{liu2023grounding}  & SWIN-T & Objects365,GoldG,Cap4M  & 12.05	& 17.29	& 15.45 &	3.72\\
                        \quad GLIP \cite{li2021grounded}  & SWIN-L & FourODs,GoldG,Cap24M  & 17.01	&23.36	&19.86	&8.40\\
                        \quad MQ-GLIP-Text \cite{xu2024multi}  & SWIN-L & Objects365,FourODs,GoldG,Cap24M &17.01	&23.36	&19.85	&8.41\\
                    \midrule
                    \midrule
                    \textbf{Prompt Engineering} & & &  & & & \\
                    \midrule
                        \quad Detic \cite{zhou2022detecting}  & SWIN-B & LVIS, COCO, IN-21K &14.92	&26.48	&17.29	&2.53\\
                        \quad GLIP \cite{li2021grounded}  & SWIN-L & FourODs,GoldG,Cap24M &17.15	&23.82	&19.36	&9.02\\
                    \midrule
                    \textbf{Standard Fine-Tuning} & & &  & & & \\
                    \midrule
                        \quad RegionCLIP \cite{zhong2022regionclip}  & RN50 & CC3M &3.84	&6.13	&5.07	&0.49\\
                        \quad Detic \cite{zhou2022detecting}  & SWIN-B & LVIS, COCO, IN-21K &15.12	&22.74	&18.99	&4.25\\
                    \midrule
                    \textbf{Federated Fine-Tuning (Ours)} & &  & & & &\\
                    \midrule
                        \quad Detic \cite {zhou2022detecting}  & SWIN-B & LVIS, COCO, IN-21K &16.58	&27.12	&19.71	&4.13\\
                        \quad Detic \cite {zhou2022detecting} w/ Prompt Engineering  & SWIN-B & LVIS, COCO, IN-21K &16.96	&27.89	&19.94	&4.37\\
                    \midrule
                    \textbf{Language Prompt Tuning} & & &  & & &\\
                    \midrule
                        \quad GLIP \cite{li2021grounded}  & SWIN-L & FourODs,GoldG,Cap24M &17.79	&21.07	&22.87	&9.12\\
                    \midrule
                    \textbf{Visual Prompting} & & &  & & & \\
                    \midrule
                        \quad MQ-GLIP-Image \cite{xu2024multi}  & SWIN-L & Objects365,FourODs,GoldG,Cap24M &13.42	&23.05	&15.00	&3.54\\
                    \midrule
                    \textbf{Multi-Modal Prompting} & &  & & & & \\
                    \midrule
                        \quad MQ-GLIP \cite{xu2024multi}  & SWIN-L & Objects365,FourODs,GoldG,Cap24M &\textbf{21.45}	&\textbf{32.23}	&\textbf{23.31}	&\textbf{10.30}\\
                    \midrule
                    \textbf{Multi-Modal Chat Assistants} & & &  & & &\\
                    \midrule
                    \quad GPT-4o Zero-Shot Classification \cite{achiam2023gpt} & \textit{Private} & \textit{Private} &9.95	&16.81	&12.11	&1.71\\
 
                    \bottomrule
                    
                \end{tabular}
            }
            \label{tab:all_res_5_shots}
        \end{table}

\begin{table}[t]
            \centering
            \caption{\small
            \small {\bf Empirical Analysis of Baselines (30-shot) on nuImages}. 
            }
            \vspace{-2mm}
            \def\arraystretch{1.0}%
            \resizebox{0.99\linewidth}{!}{
            
                \begin{tabular}{@{}l@{\ \ \ \ \ \ \ \ \ \ \ \ \ \ \ }c@{\ \ \ \ \ \ \ \ \ }c@{\ \ \ \ \ \ \ \ \ \ \ \ \ }c@{\ \ \ \ \ \ \ \ \ }c@{\ \ \ \ \ \ \ \ \ }c@{\ \ \ \ \ \ \ \ \ }c@{}}
                    \toprule
                    \multirow{2}{*}{Approach} & \multirow{2}{*}{Backbone} & 
                    Pre-Train & \multicolumn{4}{c}{Average Precision (AP)}  \\
                      & & Data 
                                        & {\tt All} & {\tt Many}    & {\tt Med}   & {\tt Few}    \\ 
                   
                    \midrule
                        \textbf{Zero-Shot Detection} & & & & & & \\
                    \midrule
                        \quad RegionCLIP \cite{zhong2022regionclip}  & RN50 & CC3M & 2.50 &	3.20 &	3.80 &	0.40 \\
                        \quad Detic \cite{zhou2022detecting}  & SWIN-B & LVIS, COCO, IN-21K & 14.40	&25.83	&16.59	&2.32  \\
                        \quad GroundingDINO \cite{liu2023grounding}  & SWIN-T & Objects365,GoldG,Cap4M  & 12.05	& 17.29	& 15.45 &	3.72\\
                        \quad GLIP \cite{li2021grounded}  & SWIN-L & FourODs,GoldG,Cap24M  & 17.01	&23.36	&19.86	&8.40\\
                        \quad MQ-GLIP-Text \cite{xu2024multi}  & SWIN-L & Objects365,FourODs,GoldG,Cap24M &17.01	&23.36	&19.85	&8.41\\
                    \midrule
                    \midrule
                    \textbf{Prompt Engineering} & & &  & & & \\
                    \midrule
                        \quad Detic \cite{zhou2022detecting}  & SWIN-B & LVIS, COCO, IN-21K &14.92	&26.48	&17.29	&2.53\\
                        \quad GLIP \cite{li2021grounded}  & SWIN-L & FourODs,GoldG,Cap24M &17.15	&23.82	&19.36	&9.02\\
                    \midrule
                    \textbf{Standard Fine-Tuning} & & &  & & & \\
                    \midrule
                        \quad RegionCLIP \cite{zhong2022regionclip}  & RN50 & CC3M &3.87	&6.05	&5.14	&0.57\\
                        \quad Detic \cite{zhou2022detecting}  & SWIN-B & LVIS, COCO, IN-21K &17.22	&25.98	&21.64	&4.78\\
                    \midrule
                    \textbf{Federated Fine-Tuning (Ours)} & &  & & & &\\
                    \midrule
                        \quad Detic \cite {zhou2022detecting}  & SWIN-B & LVIS, COCO, IN-21K &18.64	&29.13	&22.44	&5.46\\
                        \quad Detic \cite {zhou2022detecting} w/ Prompt Engineering  & SWIN-B & LVIS, COCO, IN-21K &18.67	&29.13	&22.43	&5.57\\
                    \midrule
                    \textbf{Language Prompt Tuning} & & &  & & &\\
                    \midrule
                        \quad GLIP \cite{li2021grounded}  & SWIN-L & FourODs,GoldG,Cap24M &20.73	&24.95	&\textbf{25.60}	&\textbf{11.54}\\
                    \midrule
                    \textbf{Visual Prompting} & & &  & & & \\
                    \midrule
                        \quad MQ-GLIP-Image \cite{xu2024multi}  & SWIN-L & Objects365,FourODs,GoldG,Cap24M &14.26	&24.55	&16.73	&2.79\\
                    \midrule
                    \textbf{Multi-Modal Prompting} & &  & & & & \\
                    \midrule
                        \quad MQ-GLIP \cite{xu2024multi}  & SWIN-L & Objects365,FourODs,GoldG,Cap24M &\textbf{21.40}	&\textbf{32.08}	&23.31	&10.27\\
                    \midrule
                    \textbf{Multi-Modal Chat Assistants} & & &  & & &\\
                    \midrule
                    \quad GPT-4o Zero-Shot Classification \cite{achiam2023gpt} & \textit{Private} & \textit{Private} &9.95	&16.81	&12.11	&1.71\\
                    \bottomrule
                    
                \end{tabular}
            }
            \vspace{-2mm}
            \label{tab:all_res_30_shots}
        \end{table}

\section{Empirical Analysis of Baselines (5-shot and 30-shot)}
We evaluate all baselines for the nuImages experiments with 5-shot and 30-shot in Table~\ref{tab:all_res_5_shots} and \ref{tab:all_res_30_shots}, respectively. We find that trends from the main paper hold. Notably, MQ-GLIP with-multi-modal prompting performs the best. However, we find that adding more examples (e.g. MQ-GLIP 5-shot vs. MQ-GLIP 30-shot) does not seem to help in-context learning based methods nearly as much as gradient-based fine-tuning approaches. 

\section{Foundational FSOD with LVIS}
    \begin{table}[t]
            \centering
            \caption{\small
            \textbf{LVIS Foundational FSOD Performance}. We present fine-tuning results for different variants of Detic on the LVIS 10-shot dataset. We follow the standard FSOD setup and pre-train Detic on {\tt LVIS-base} for fair comparison with prior work. Detic pre-trained only on {\tt LVIS-base} outperforms specialized methods like TFA and DiGeo by $\sim$6 AP, {\em without fine-tuning} on rare classes. Since we keep the model backbone (ResNet-50) and pre-training data same for all methods, these performance improvements can be attributed to Detic's CLIP-based classifier. This demonstrates that concept leakage through language significantly improve FSOD, and leveraging language cues should be embraced in data constrained settings. Naively fine-tuning Detic yields a performance drop of $AP_f$ and $AP_c$ because treating common classes as negatives in rare category federated datasets hurts performance. Instead, we find that embracing the federated nature of FSOD datasets provides consistent improvements in fine-tuning (30.0 vs. 30.8 for ResNet-50). Further, pseudo-labeling negatives in each image provides a modest improvement (30.8 vs. 31.6 for ResNet-50). Similar trends hold for the SWIN-B and SWIN-L backbones.}
            \resizebox{0.9\linewidth}{!}{
                \begin{tabular}{@{}l@{\ \ \ \ \ \ \ \ \ \ \ \ \ \ \ \ \ \ \ \ \ \ \ }c@{\ \ \ \ \ \ \ \ \ \ \ \ \ \ \ }c@{\ \ \ \ \ \ \ \ \ \ \ \ \ \ \ }c@{\ \ \ \ \ \ \ \ \ \ \ \ \ \ \ }c@{}}
                    \toprule
                    \multirow{2}{*}{Approach} & \multicolumn{4}{c}{10-shots}  \\
                                                                                    & $AP$  & $AP_f$ & $AP_c$ & $AP_r$   \\ 
                    \midrule
                    \textbf{ResNet-50 Backbone} & & & & \\
                      \midrule
                        TFA w/ fc~\cite{wang2020frustratingly}                     & 24.1  & 27.9  & 23.9 & 14.9  \\
                        TFA w/ cos~\cite{wang2020frustratingly}                    & 24.4  & 27.7  & 24.3 & 16.9  \\
                        DiGeo~\cite{ma2023digeo}                                   & 24.9  & 28.5  & 24.6 & 17.3  \\
                    \midrule
                    Detic ({\tt Base} Only) \cite{zhou2022detecting}                                      & 30.0  & 34.4 & 30.8 & 16.3  \\
                        \quad + Fine-Tuning  ({\tt Base} + {\tt Novel})                                  & 30.0  & 33.2 & 31.9 & 15.5  \\
                        \quad w/ FedLoss  & 30.8  & 33.9 & 32.7 & 17.4  \\
                        \quad w/ Pseudo-Negatives        & \textbf{31.6}  & \textbf{34.8} &   \textbf{32.8} & \textbf{19.8}   \\
                    \midrule
                    \textbf{Swin Backbone} & & & & \\
                    \midrule 
                        Detic ({\tt Base} Only, SWIN-B) \cite{zhou2022detecting}                                      & 35.2  & 38.7 & 36.8 & 21.4  \\
                        \quad + Fine-Tuning  ({\tt Base} + {\tt Novel})                                  & 35.9  & 37.1 & 37.8 & 26.7  \\
                        \quad w/ FedLoss  & 36.5  & 36.7 & 38.3 & 30.4  \\

                        \quad w/ Pseudo-Negatives         &  37.2 &  37.7 & 38.2 & 32.6    \\
                        \midrule
                        MQ-GLIP-Text \textit{(SWIN-L)} & 35.8 & 40.2& 33.1 & 33.0\\
                        MQ-GLIP-Image \textit{(SWIN-L)} & 28.8 & 33.0 & 26.6 & 25.1 \\
                        MQ-GLIP \textit{(SWIN-L)} & \textbf{43.4} & \textbf{46.4} & \textbf{41.8} & \textbf{40.1}\\ 

                    \bottomrule
                \end{tabular}
                }
        \label{tab:lvis_v05_supp}
            \vspace{-4mm}
    \end{table}

Although we use nuImages for Foundational FSOD for benchmarking in the main paper and in our competition, other datasets can still be evaluated under this framework. We include benchmarking results for LVIS below. LVIS~\cite{gupta2019lvis} re-annotates COCO images using 1,230 fine-grained classes, which are divided into frequent, common and rare based on the cardinality of each class. Frequent and common classes are combined to form {\tt LVIS-base} and is used for pre-training. Rare classes are used for {\tt LVIS-novel}. Following \cite{wang2020frustratingly, ma2023digeo}, we benchmark with LVIS v0.5 on publicly released data splits and report performance averaged across 3 splits for frequent, common, and rare groups ($AP_f, AP_c, AP_r$) on the LVIS val-set. 

As shown in Table~\ref{tab:lvis_v05_supp}, Detic outperforms all recent FSOD baselines including DiGeo~\cite{ma2023digeo} by $\sim$6 $AP_c$ \& $AP_f$  and achieves $16.3$ $AP_r$ without ever seeing any rare class data (e.g., by prompting Detic ({\tt Base} Only) with the rare class names). Importantly, these performance improvements can be attributed to Detic's CLIP-based classifier, which uses CLIP text embeddings corresponding to class names. Such embeddings are a result of large-scale pre-training, which we can effectively leverage for the few-shot task. This highlights the role of language in data-constrained settings. 

Further, fine-tuning Detic with pseudo-negatives improves overall performance by $1.6$ AP ($30.0$ vs $31.6$) over naive fine-tuning. To contextualize the improvement in performance, we note that between TFA (ICML 2020) and DiGeo (CVPR
2023), the community improved on LVIS FSOD by only $0.5$ AP (cf. Table \ref{tab:lvis_v05_supp}). Finally, we note that simply replacing the ResNet-50 backbone with a Swin-B transformer yields a sizeable $12.8$ AP improvement for rare classes ($19.8$ vs. $32.6$).

We present fine-tuning results for different variants of Detic on the LVIS 10-shot dataset. Following the standard FSOD protocol, we pre-train Detic on {\tt LVIS-base} (e.g. frequent and common classes) and fine-tune on 10-shots from each class in {\tt LVIS-base} and {\tt LVIS-novel}. Importantly, this means that only results for $AP_r$ are indicative of true few-shot performance. First, we find that naively fine-tuning Detic on {\tt Base + Novel} yields lower performance for $AP_f$ and $AP_r$. Intuitively, this suggests that ignoring the federated nature of FSOD datasets (e.g. by following the standard practice of assuming common classes are negatives for rare class federated datasets) hurts common class performance (cf. Table \ref{tab:lvis_v05_supp}). Importantly, simply training with FedLoss significantly improves over naive fine-tuning, increasing $AP_r$ by 1.9\% (15.5 vs. 17.4) and 3.7\% (26.7 vs. 30.4) for the ResNet-50 and Swin backbones respectively. Further, leveraging our proposed negative pseudo-labeling strategy provides further improvements over the naive federated loss, increasing $AP_r$ by another 2.4\% (17.4 vs. 19.8) and 3.7\% (30.4 vs. 32.6) for the ResNet-50 and Swin backbones respectively. Similar to nuImages, we find that multi-modal prompting with MQ-GLIP performs the best of all baselines tested, significantly improving over MQ-GLIP-Text and MQ-GLIP-Image. We attribute MQ-GLIP's strong performance to its bigger backbone and significantly larger pre-training dataset.

\textbf{LVIS v0.5 Detic Experiment Details.}
We select Detic with a Resnet-50 backbone for fair comparison with prior work. We pre-train Detic on {\tt LVIS-base} for $90k$ iterations with a batch size of $32$ using an AdamW optimizer and a learning rate of $2e$-3. All images are resized to  $640 \times 640$ and we also enable Repeat Factor Sampling~\cite{gupta2019lvis}. Following \cite{wang2020frustratingly}, we sample {\em up to} $10$ shots for each class in LVIS (since all classes may not have 10 examples). We use a batch size of $32$, learning rate of $2.5e$-5 for $46k$ iterations. We do not use Repeat Factor Sampling for fine-tuning. We sample $50$ categories for each training image, i.e., $|S|=50$ for the FedLoss experiments. We derive negatives from pseudolabels with at least $20\%$ confidence for the Psuedo-Negative experiment.

\newpage
\section*{Checklist}


\begin{enumerate}

\item For all authors...
\begin{enumerate}
  \item Do the main claims made in the abstract and introduction accurately reflect the paper's contributions and scope?
    \answerYes{}
  \item Did you describe the limitations of your work?
    \answerYes{} A separate limitations sections is present in the main paper
  \item Did you discuss any potential negative societal impacts of your work?
    \answerNA{}
  \item Have you read the ethics review guidelines and ensured that your paper conforms to them?
    \answerYes{}
\end{enumerate}

\item If you are including theoretical results...
\begin{enumerate}
  \item Did you state the full set of assumptions of all theoretical results?
    \answerNA{}
	\item Did you include complete proofs of all theoretical results?
    \answerNA{}
\end{enumerate}

\item If you ran experiments (e.g. for benchmarks)...
\begin{enumerate}
  \item Did you include the code, data, and instructions needed to reproduce the main experimental results (either in the supplemental material or as a URL)?
    \answerYes{} See Supplement
  \item Did you specify all the training details (e.g., data splits, hyperparameters, how they were chosen)?
    \answerYes{} See Supplement
	\item Did you report error bars (e.g., with respect to the random seed after running experiments multiple times)?
    \answerYes{} See Supplement
	\item Did you include the total amount of compute and the type of resources used (e.g., type of GPUs, internal cluster, or cloud provider)?
    \answerYes{} See Supplement
\end{enumerate}

\item If you are using existing assets (e.g., code, data, models) or curating/releasing new assets...
\begin{enumerate}
  \item If your work uses existing assets, did you cite the creators?
    \answerYes{}
  \item Did you mention the license of the assets?
    \answerYes{}
  \item Did you include any new assets either in the supplemental material or as a URL?
    \answerYes{} Our code and data are the new assets, and are part of the supplement
  \item Did you discuss whether and how consent was obtained from people whose data you're using/curating?
    \answerNA{}
  \item Did you discuss whether the data you are using/curating contains personally identifiable information or offensive content?
    \answerNA{}
\end{enumerate}

\item If you used crowdsourcing or conducted research with human subjects...
\begin{enumerate}
  \item Did you include the full text of instructions given to participants and screenshots, if applicable?
    \answerNA{}
  \item Did you describe any potential participant risks, with links to Institutional Review Board (IRB) approvals, if applicable?
    \answerNA{}
  \item Did you include the estimated hourly wage paid to participants and the total amount spent on participant compensation?
    \answerNA{}
\end{enumerate}

\end{enumerate}

\end{document}